\documentclass[journal]{IEEEtran}
\usepackage[utf8]{inputenc}

\usepackage{amsmath}
\usepackage{amssymb}
\usepackage{amsfonts}
\usepackage{graphicx}
\usepackage{subfig}
\usepackage{booktabs}
\usepackage{afterpage}
\usepackage{hyperref}

\newcommand{\norm}[1]{\left\lVert#1\right\rVert}

\DeclareMathOperator*{\argmax}{arg\!max}
\graphicspath{{images/},{images_extra/}}

\begin{document}

\title{Characterizing and evaluating adversarial examples for Offline Handwritten Signature Verification}

\author{Luiz~G.~Hafemann,
	Robert Sabourin,~\IEEEmembership{Member,~IEEE,}
	and~Luiz~S.~Oliveira
	\thanks{L. G. Hafemann and R. Sabourin are with the Laboratoire d'imagerie, de vision et d'intelligence artificielle, \'Ecole de technologie sup\'erieure, Universit\'e du Qu\'ebec, Montreal, Canada. (e-mail: lghafemann@livia.etsmtl.ca, robert.sabourin@etsmtl.ca)}
	\thanks{L. S. Oliveira is with the Department of Informatics, Federal University of Parana, Curitiba, Brazil (e-mail:lesoliveira@inf.ufpr.br)}.
	\thanks{This work was supported by the Fonds de recherche du Qu\'ebec - Nature et technologies (FRQNT), the CNPq grant \#206318/2014-6 and by the grant RGPIN-2015-04490 to Robert Sabourin from the NSERC of Canada.}
	}

\IEEEpubid{1556-6013 \copyright 2019 IEEE. Personal use is permitted, but republication/redistribution  requires  IEEE permission. }

\maketitle

\begin{abstract}

The phenomenon of Adversarial Examples is attracting increasing interest from the Machine Learning community, due to its significant impact to the security of Machine Learning systems. Adversarial examples are similar (from a perceptual notion of similarity) to samples from the data distribution, that ``fool" a machine learning classifier. For computer vision applications, these are images with carefully crafted but almost imperceptible changes, that are misclassified. In this work, we characterize this phenomenon under an existing taxonomy of threats to biometric systems, in particular identifying new attacks for Offline Handwritten Signature Verification systems. We conducted an extensive set of experiments on four widely used datasets: MCYT-75, CEDAR, GPDS-160 and the Brazilian PUC-PR, considering both a CNN-based system and a system using a handcrafted feature extractor (CLBP). We found that attacks that aim to get a genuine signature rejected are easy to generate, even in a limited knowledge scenario, where the attacker does not have access to the trained classifier nor the signatures used for training. Attacks that get a forgery to be accepted are harder to produce, and often require a higher level of noise - in most cases, no longer ``imperceptible" as previous findings in object recognition. We also evaluated the impact of two countermeasures on the success rate of the attacks and the amount of noise required for generating successful attacks.

\end{abstract}

\begin{IEEEkeywords}
	Adversarial Machine Learning, Signature Verification, Biometrics
\end{IEEEkeywords}

\section{Introduction}

Biometric systems are extensively used to establish a person's identity in legal and administrative tasks  \cite{jain_introduction_2004}. They are commonly modeled as Pattern Recognition systems, in which biometric data from an individual is acquired (e.g. during an enrollment process), and stored as a ``template" for future comparisons, or used to train a classifier that can discriminate if new samples belong to this user. 

The reliability of these systems have security implications, and in the last decade these systems have been analyzed from an Adversarial Machine Learning perspective. From this viewpoint, we consider an active adversary, with its own goals (e.g. getting access to a system), knowledge (e.g. knowing the classifier parameters, or the learning algorithm) and capabilities (e.g. ability to manipulate the training data, or the inputs during test time). In particular, Ratha et al. \cite{ratha_analysis_2001} and later Biggio et al. \cite{biggio_adversarial_2015} characterize the different components of a biometric system that can be attacked.

However, an emerging issue of ``Adversarial Examples" pose new security concerns for such systems. This issue refers to adversarial input perturbations specially crafted to induce misclassifications. Szegedy et al. \cite{szegedy_intriguing_2013} showed that very small perturbations on images (almost imperceptible) could be crafted to mislead a state-of-the-art CNN-based classifier. Moreover, attacks crafted for one model often transfer to other models, meaning that an attacker could train its own surrogate classifier to generate attacks, as long as it has access to data from the same data distribution. This issue has been analyzed in many recent papers \cite{goodfellow_explaining_2014,papernot_distillation_2016,tramer_ensemble_2017,carlini_towards_2017, carlini_adversarial_2017}, but the theoretical reasons are not fully understood, and most defenses are weak (i.e. they fail if the attacker knows about the defense).

We evaluate this new threat for biometric systems, by characterizing the potential new attacks under a taxonomy of threats to such systems \cite{ratha_analysis_2001}, \cite{biggio_adversarial_2015}. We consider particular attack scenarios to Offline Handwritten Signature Verification, identifying the attacker's goals, required knowledge and capabilities. 

It is worth noting that attacking verification systems can present difficulties not present in classification problems. In particular, as new users join the system, they introduce a new \emph{class}, not only unseen examples of existing classes. We present a refined version of the adversary's knowledge model that explicitly makes the distinction of whether access to data from a particular individual of interest is available to the attacker. 

\begin{figure*}
	\centering
	\includegraphics[width=0.7\textwidth]{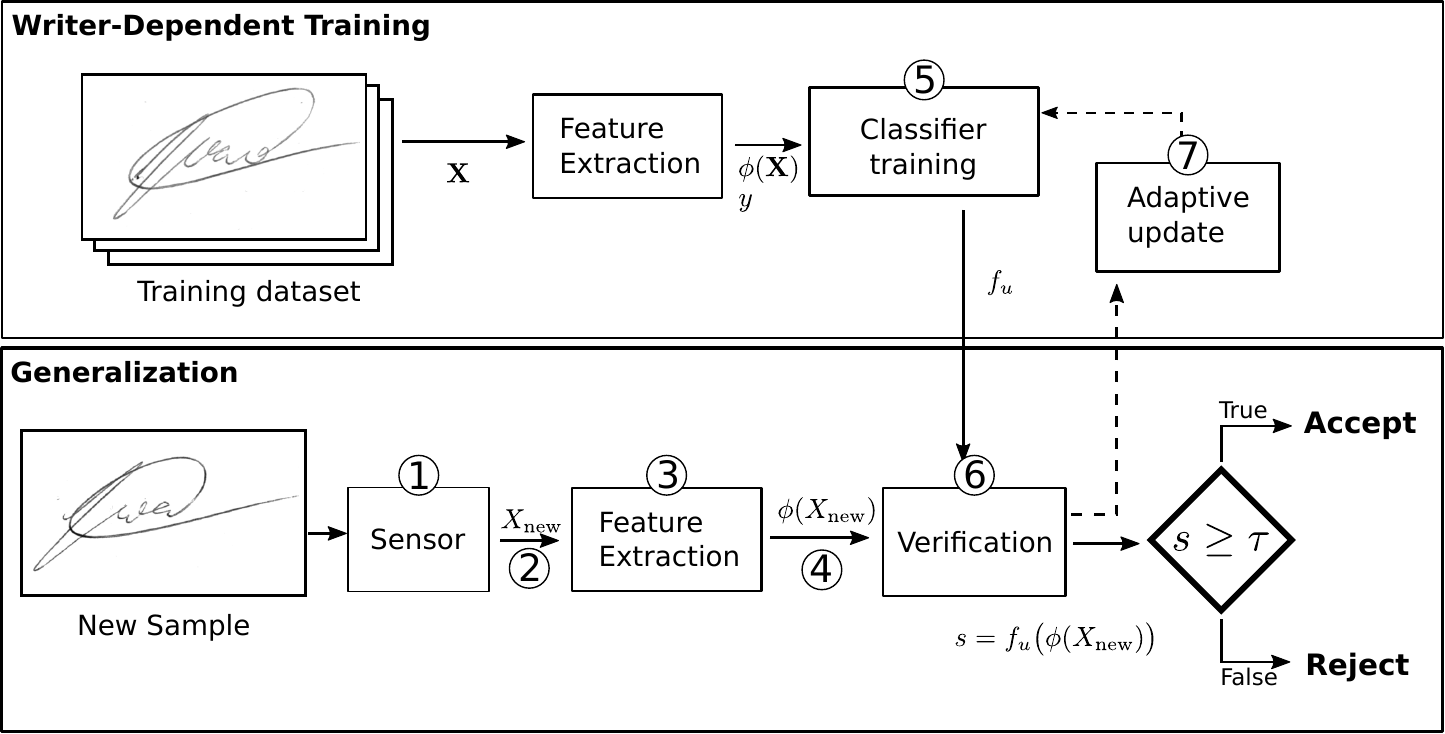}
	\caption{A typical writer-dependent signature verification system, with annotated points of attack. On the training phase, a classifier $f_u$ is trained for each user. During operations, for a new sample $X_\text{new}$ we obtain a feature vector $\phi(X_\text{new}$), and use the classifier $f_u$ to accept or reject the signature. For adaptive systems, an update rule select signatures for classifier adaptation.}
	\label{fig:attack_surface}
\end{figure*}

We conducted experiments on Writer-Dependent classifiers trained with a CNN-based representation (SigNet) and a handcrafted feature extractor (CLBP), considering four widely used Datasets: MCYT, CEDAR, GPDS-160 and the Brazilian PUC-PR. We defined a comprehensive set of experiments to evaluate such systems under different scenarios of the adversary's knowledge level and objectives, using four attack methods (gradient-based and gradient-free). Our main contributions are as follows:

\IEEEpubidadjcol

\begin{itemize}
	\item We characterize different attack scenarios for Offline Handwritten Signature Verification systems, focused on new threats introduced by Adversarial Examples.
	\item We identify that there is an asymmetry in the attacks, empirically showing that attacking genuine signatures (so that they are rejected) can be done with high success rate and a relatively low amount of noise, while attacking forgeries (so that they are accepted) is a much harder. 
	\item Our experiments with different scenarios of attacker knowledge show that attacks can be done even with Limited Knowledge, where the attacker has no access to the signatures used to train the classifiers, showing that this transferability affects both CNN-based systems and systems based on handcrafted features. We also identify that attack transferability is greatly reduced if the CNN is trained on a different subset of \emph{users}, contrasting with previous findings that attacks transfer well if the CNN is trained on a different subset of samples from the same classes \cite{szegedy_intriguing_2013}.
	\item Lastly, we evaluate the impact of countermeasures and find that the Madry defense \cite{madry_towards_2017} is effective in increasing the amount of noise necessary to make a sample adversarial, even when it is applied only to the feature learning phase, and not on training the WD classifiers. Code for reproducing the experiments will be made publicly available at \url{https://github.com/luizgh/adversarial_signatures}.	
\end{itemize}

The paper is organized as follows: in section \ref{sec:security_biometric} we introduce the main concepts of security in biometric systems; in section \ref{sec:adversarial_examples} we present the issue of adversarial examples and in section \ref{sec:attack_scenarios} we present particular attack scenarios for offline signature verification, and a refinement of the adversary's knowledge model. Section \ref{sec:experimental} describes the experimental protocol, and the results are discussed in section \ref{sec:results}. Finally, our conclusions are listed in section \ref{sec:conclusion}.

\section{Security in biometric systems}
\label{sec:security_biometric}

The security of machine learning systems have been widely studied in the past decade. Barreno et al. \cite{barreno_can_2006, barreno_security_2010} categorize attacks to such systems along three axes: (i) the influence of the attack, that can be causative (when training data is compromised) or exploratory (probing the learner to acquire information); (ii) the specificity of the attack: \emph{targeted}, in which a particular point or a set of points is targeted or \emph{indiscriminate}; and (iii) the security violation of the attack, that can seek an integrity violation (e.g. intrusion) or availability disruption (e.g. make the system unusable for legitimate users).

Biggio et al. \cite{biggio_adversarial_2015, biggio_security_2014} further expands this analysis for biometric systems, incorporating a model of the adversary that includes its \emph{goals}, \emph{knowledge} of the target system, and \emph{capabilities} of manipulating the input data or system components. The goals of an attacker are mainly divided in: 1) \emph{Denial of service}: preventing real users from using the system; 2) \emph{Intrusion}: impersonating another user; 3) \emph{Privacy violation}: stealing private information from an user (such as the biometric templates). The \emph{knowledge} of the adversary refers to the information of the target system that is available to the adversary, such as perfect knowledge (e.g. knowledge of the feature extractor, type of classifier and model parameters) or limited (partial) knowledge of the system. The \emph{capabilities} of the adversary refer to what it can \emph{change} in the target system, such as changing the training set (poisoning attack), or the inputs to the system at test time (evasion attack).

Modeling the knowledge of the adversary was formalized by Biggio and Roli \cite{biggio_wild_2017}. Let $\mathcal{X}$ and $\mathcal{Y}$ be the feature and label spaces, respectively, and $\mathcal{D}$ be a dataset $\mathcal{D} = \{x_i, y_i\}_{i=1}^n$ of $n$ training samples. Let $f$ be a training algorithm (classifier), and $w$ be a collection of its parameters and hyper-parameters. The knowledge of the attacker can be formalized as a set $\theta$, containing the components of the system that are known to the attacker. Perfect-Knowledge (PK) attacks consider full knowledge of the system, that is, $\theta_{PK} = (\mathcal{D}, \mathcal{X}, f, w)$. We can also consider Limited Knowledge (LK) attacks, in which some of the information is not available to the adversary. As an example, if the adversary does not have access to the learned weights of the model, but has access to the training data, a surrogate classifier $f$ can be trained (learning parameters $\hat{w}$) and used to generate the attack. Similarly, if the training data is not available, the adversary may be able to collect another training set from the same data distribution and use it to train the surrogate classifier. In this last scenario, the knowledge of the attacker would be represented as $\theta_{LK} = (\mathcal{\hat{D}}, \mathcal{X}, f, \hat{w})$. The hat symbol ($\hat{~}$) indicate limited knowledge of a component (such as getting a surrogate dataset from the same data distribution).

Biometric systems are composed of several components, such as the sensors capturing the biometric, and software to extract features, store templates and perform classification. Ratha et al \cite{ratha_analysis_2001} identified eight points of attack on biometric security systems, that were later grouped by Jain et al \cite{jain_biometric_2008} and extended by Biggio et al \cite{biggio_adversarial_2015} to include multi-modal systems and adaptive systems. The set of this attack points is considered the \emph{attack surface} of the system. Figure \ref{fig:attack_surface} shows a typical User-Dependent classification system, with the main attack points. Below we discuss the main threats to the different points of attack.

The first point of attack (\#1) in a biometric system is the user interface that collects the sample (e.g. a scanner capturing a document with a signature, or a mobile application taking a picture of a bank cheque). For many biometrics, attacks on this first point mainly consist of spoofing attacks, that normally use a fabricated fake biometric trait. Possible defenses for such attacks rely on liveness detection. On the signature verification task, simulated and traced forgeries can be considered attacks targeting this stage. A second set of attack points refer to attacks in the communication between different components of the system (\#2, \#4) (for example, intercepting and replacing the sensor input or the extracted features, that is input to the subsequent module). Defenses for such attacks involve encrypting the communication between the different modules. The software modules (\#1, \#3, \#5, \#6, \#7) may present vulnerabilities in the code (such as buffer overflow) that can be exploited by a malicious user. The classifier training (\#5) can be targeted for poisoning attacks (e.g. adding samples from another user in the training data for subsequent intrusion). For adaptive systems, the template update rule (\#7) can be targeted to update the template database (e.g. for intrusion).

\section{Adversarial Examples}
\label{sec:adversarial_examples}

\begin{figure}
	\centering
	\includegraphics[width=\columnwidth]{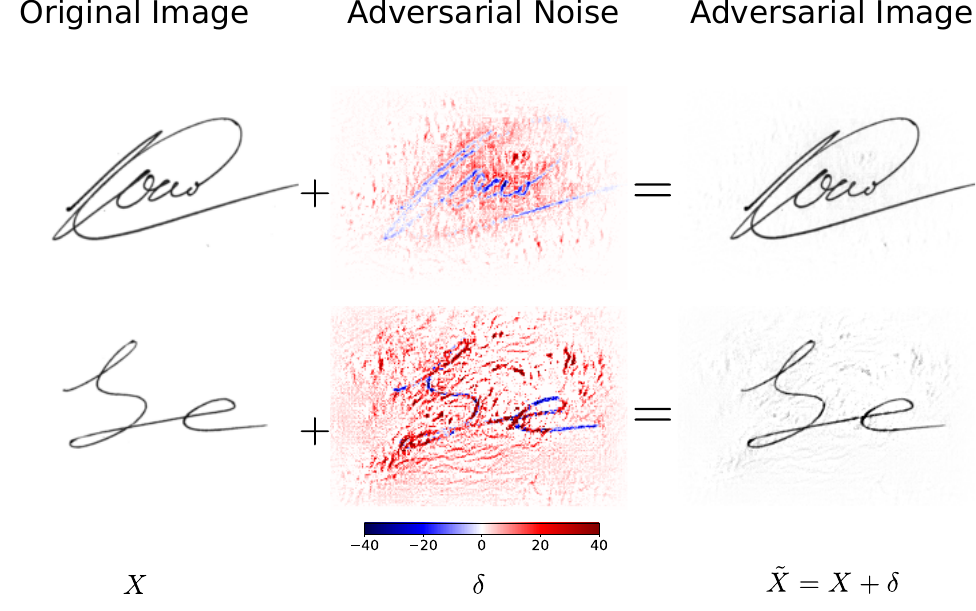}
	\caption{Illustration of adversarial examples. An adversarial noise $\delta$ is added to original images $X$, such that the resulting image $\tilde{X}$ is misclassified. \textbf{Top}: Type-I attack: a genuine signature from user $u_1$ (left) is attacked to be classified as a forgery (right). \textbf{Bottom}: Type-II attack. The original image (left) is from user $u_2$ (i.e. a random forgery for $u_1$), and is attacked to be classified as a genuine (right).}
	\label{fig:adv_example} 
\end{figure}

Adversarial examples are samples similar to the true data distribution, but that fool a classifier. In computer vision, these are images $\tilde{X}$ that are visually similar to a ``real" image $X$, but that fool a classifier (i.e. the classifier predicts an incorrect class for $\tilde{X}$: $\argmax_y {P(y | \tilde{X})} \neq \argmax_y {P(y | X)}$). 

Szegedy et al. \cite{szegedy_intriguing_2013} showed that for deep neural networks, we can run an optimization procedure to produce a small change $\delta$ to an image, such that $\tilde{X} = X + \delta$ is an adversarial example, as illustrated in Figure \ref{fig:adv_example}. Perhaps more surprisingly, they also discovered that an attack that is created to fool one network also fools other networks (trained on different subsets of data), meaning that attacks can be created even without full knowledge of the classifier under attack. It was later shown that such attacks can be done in the physical world \cite{kurakin_adversarial_2016}, where adversarial images printed on paper and later captured with a camera also fooled a classifier. Lastly, although some defense strategies have been proposed \cite{goodfellow_explaining_2014,papernot_distillation_2016,tramer_ensemble_2017,madry_towards_2017}, most solutions are not robust to strong iterative attacks. Even \emph{detecting} that an input is adversarial is a hard task \cite{carlini_adversarial_2017}. 

Most of the recent research on this area concentrates on differentiable classifiers (usually Deep Learning models), creating attacks that use gradient information of the loss function with respect to the inputs. However, most feature extractors used in signature verification (such as LBP, HOG) are non-differentiable, and therefore attacks to systems using these features could not rely on gradient-based methods. Some methods proposed in the literature do not rely on gradient information, and could potentially be used for this task. Papernot et al. proposed \emph{Substitute model training} \cite{papernot_practical_2017}, in which the attacker train a substitute (differentiable) model, and use it to generate the attack.
Brendel et al. proposed a \emph{Decision-based attack} \cite{brendel_decision-based_2017}, that relies only on the decision (prediction) of the model under attack. Its strategy is the opposite of most attacks: given an image $X$ and an image $\tilde{X}^0$ that is from another class, the algorithm iteratively refines $\tilde{X}^k$ to be closer to $\tilde{X}$ (e.g. in $L_2$ norm). The image $\tilde{X}^0$ can be a random image (e.g. sampled at random until it is from the desired class), or an actual image from a target class. Chen et al. proposed a  \emph{Zeroth order optimization} method \cite{chen_zoo:_2017}, where the gradient is estimated numerically. Doing so naively is impractical (due to the dimensionality of the input), so the authors employ techniques to reduce the computational complexity of this estimation (block coordinate descent, attack-space dimension reduction, hierarchical attacks and importance sampling). With all these techniques, the attack has shown to scale to imagenet ($299 \times 299 \times 3$ pixels), producing an attack in 20 minutes. This method requires the function to be smooth and Lipshitz continuous. Ilyas et al. proposed using \emph{Natural Evolution Strategy (NES) gradient estimate} \cite{ilyas_query-efficient_2017} - instead of using numerical methods to estimate the gradient (as above), the authors propose using Natural Evolution Strategies for the gradient estimate. These estimates are given by computing the loss function along random directions. The authors claim that this method require 1-2 orders of magnitude less computations of the loss function. Lastly, Ramanathan et al. \cite{ramanathan_adversarial_2017} explored using \emph{Simulated annealing} for creating adversarial examples for a system based on HOG features with a linear SVM classifier. In each iteration, a small perturbation is applied to the image, and the distance of the new image to the SVM hyperplane is used as a condition to accept the new point. With this approach the authors were able to craft adversarial images with imperceptible noise that fooled the HOG+SVM classifier.

\subsection{Attacks considered in this paper}
\label{sec:attacks_considered}

In this paper, we consider two gradient-based attacks (that can be used when the classifiers are differentiable with respect to the input), and two gradient-free attacks, that can be used even if the features and/or classifiers are non-differentiable. In this paper we are mostly interested in feature extractors widely used for signature verification, and chose the LBP descriptor, which is used in several studies \cite{vargas_off-line_2011,hu_offline_2013,yilmaz_score_2016}. Since LBP is highly discontinuous (due to the thresholding using the center pixel's value), methods that estimate the gradient are less interesting: the gradient should be very discontinuous (0 almost everywhere), since for each pixel, the transition from one pattern to the other is a step function. For this reason we selected two methods that do not rely on estimating the gradients: the decision-based attack \cite{brendel_decision-based_2017} and the optimization using Simulated annealing. For the gradient-based attacks, we considered the Fast Gradient Method (FGM) \cite{goodfellow_explaining_2014} and the Carlini \& Wagner attack \cite{carlini_towards_2017}. 

The decision-based attack \cite{brendel_decision-based_2017} is an iterative method: given an image $X$ from class $y_i$, the objective is to find an image $\tilde{X}^k$ that is classified as a different class, and minimizes the distance $D(X, \tilde{X}^k)$ for some distance measure. It starts with a sample $\tilde{X}^0$ from a class $y \neq y_i$. In each step, first the sample is projected in a random direction that is orthogonal to $(\tilde{X}^{k-1} - X)$ (i.e. orthogonal to a straight line to the sample X), and then takes a step in the direction of $X$. If the point is still from a class different than $y_i$, it is accepted as the next point $\tilde{X}^k$, otherwise a new point is searched in another random direction. This method therefore only requires the decision of the model (which class a sample $\tilde{X}^k$ belongs to).

The annealing method uses the well known simulated annealing method as a gradient-free optimization method. Starting from the image $X$, we add a small perturbation obtaining $\tilde{X}^k$. If the resulting image is closer to the decision boundary of the SVM (i.e the score decreases/increases depending on the type of attack), it is accepted as the next point. Otherwise, with a probability inversely proportional to the current step, it is still accepted as the next point. In the work from Ramanathan et al. \cite{ramanathan_adversarial_2017}, the authors consider as the objective function simply to reduce the distance to the SVM hyperplane, and stop optimization when the boundary is crossed. In our experiments, we found it necessary to include a penalty on the $L_2$ norm of the noise added to the image. This is further detailed in section \ref{sec:experimental}.

The FGM attack is a one-step gradient-based attack. In this paper we consider the version of this attack focused on the $L_2$ norm:

\begin{equation}
\label{eq:fgm}
\tilde{X} = X + \epsilon \frac{\nabla{J(x, y)}}{\norm{\nabla{J(x, y)}}_2}
\end{equation}

\noindent Where $X$ is the original image, $\nabla{J(x, y)}$ is the gradient of the loss function with respect to the input, and $\epsilon$ is a hyperparameter that controls the size of the update. The adversarial image is then clipped to the allowed range of the input (e.g. pixels between 0 and 255).

The Carlini \& Wagner $L_2$ attack uses an iterative gradient attack, using a gradient descent method (the Adam optimizer). The objective to be minimized contains two terms: a term minimizing the noise $\delta$ and a term encouraging the model to misclassify the image:

\begin{equation}
\min_w \norm{\delta}_2 + c f(X + \delta) \\
\end{equation}

\noindent Where $c$ trades-off between the two objectives, and is chosen with a binary search (the smallest $c$ that still obtains a misclassified image). Instead of enforcing hard constraints on the adversarial image (to keep pixel values between 0 and 255), the authors propose a change of variable. First, they consider images normalized between 0 and 1. Then, to enforce that $X + \delta \in [0,1]$ they consider the following change of variable:

\begin{equation}
\delta_i = \frac{1}{2} (\tanh(w_i) + 1) - x_i
\end{equation}

Since $-1 \leq \tanh(w_i)  \leq 1$, it follows that $0 \leq X_i + \delta_1 \leq 1$, satisfying the box constraints on the resulting image, but putting no constraints on the variable under optimization (w). As for the term that encourages the model to misclassify the image, they choose a term that seeks to increase the distance between the logits (pre-softmax activation) of the target class $t$ and the class with maximum prediction (other than the target class):

\begin{align}
f(X) = \max(\max_{i \neq t}(Z(X)_i) - Z(X)_t, -\kappa)
\end{align}

\noindent Where $Z(X)$ is the logit (pre-softmax activation) and  $\kappa$ is a constant that can be used to select how confident the model must be in the wrong class prediction. This loss function has no constraints, and can be solved by any gradient-based method.

\subsection{Countermeasures}

Under a paradigm of \emph{Security by design}, systems should be designed to be secure from the ground up. In the case of Machine Learning, systems should be designed explicitly considering an adversary \cite{biggio_wild_2017}. Dalvi et al \cite{dalvi_adversarial_2004} presented one of the first formulations of this problem, by considering a \emph{game} between the classifier and the adversary. They propose a solution of this game for naive bayes classifiers, considering a classifier that performs as well as possible against an optimal adversary. This has some resemblance to recent approaches proposed for adversarial examples called \emph{Adversarial Training} \cite{goodfellow_explaining_2014,tramer_ensemble_2017}, in which the training procedure is augmented with adversarial samples, with the objective of increasing robustness of the systems. 

In this work we are concerned with the new vulnerabilities introduced by adversarial changes in the input images that induce misclassifications in Signature Verification systems. In this setting, some defenses become harder to implement - for instance, Biggio et al \cite{biggio_evasion_2013} propose learning the support ($P(X)$) and incorporating this knowledge on the classifier training. Learning this support when $X$ is high dimensional (which is the case in signature images, eg. $150 \times 200$ pixels in this work) is a hard task, specially when just a few samples per user are available. The problem of working with large models and input dimensions is explored in recent work in adversarial examples for deep neural networks. For instance by \emph{Adversarial training}  \cite{goodfellow_explaining_2014}, \cite{tramer_ensemble_2017}; \emph{defensive distilation} (retraining a network with knowledge extract from a previous training) \cite{papernot_distillation_2016}; and techniques to add non-differentiable steps in the inference process (e.g. transforming the input with non-differentiable operations \cite{guo_countering_2017}). Most defenses, however, have been shown to fail when the attacker has knowledge of them. Tramer et al. \cite{tramer_ensemble_2017} showed that Adversarial training is not robust to iterative attacks on a white-box (PK) scenario; Carlini and Wagner showed that distillation is also not effective in this scenario \cite{carlini_towards_2017}. More recently, Athalye et al. showed that almost all defenses presented in recent ICLR and CVPR conferences can be bypassed \cite{athalye_obfuscated_2018}, \cite{athalye_robustness_2018}. The only exception was the work of Madry et al. \cite{madry_towards_2017}, that propose a framework that provides guarantees against attacks with a maximum $L_\infty$ norm. However, as noted in \cite{athalye_robustness_2018}, this defense is hard to scale (the authors only reported results on the CIFAR-10 dataset, which consists of small images of 32 $\times$ 32 pixels), and that resistance to $L_\infty$ attacks does not guarantee resistance to other scenarios (e.g. when the attacker is limited by a maximum $L_2$ norm of the noise). This problem therefore remains as an open research question.

In this paper we focus our attention in defenses for the CNN-based models, in particular by evaluating two defenses: Ensemble Adversarial Training \cite{tramer_ensemble_2017} and the Madry defense \cite{madry_towards_2017}. The first has demonstrated some robustness in Limited Knowledge scenarios, while the second is a proposed defense against perfect-knowledge attacks.

For the ensemble adversarial training, we first train $M$ models on the task at hand. Then we train another model with the following loss function:

 \begin{equation}
 \label{eq:ensadv}
     \tilde{J}(X, y, \theta) = \alpha J(X, y, \theta) + (1-\alpha) J(\tilde{X}, y, \theta)
 \end{equation}

\noindent Where $J(X, y , \theta)$ is the cross-entropy loss function of a sample $X$ with true label $y$, and $\tilde{X}$ is an adversarial sample generated using FGM (equation \ref{eq:fgm}) either using the model being trained, or one of the $M$ previously trained models.

The Madry defense involves a saddle point optimization problem, in which we optimize for the worst case:
\begin{equation}
 \label{eq:madry}
\begin{gathered}
 \min_\theta p(\theta) \\
 \text{where} \quad p(\theta) = \mathbb{E}_{(x,y) \sim \mathcal{D}}\big[\max_{\delta \in \mathcal{S}} J(X + \delta, y, \theta) \big]
\end{gathered}
\end{equation}

\noindent Where $\mathcal{S}$ defines a feasible region of the attack (i.e. the attacker capability). For instance, to add robustness against attacks that minimize the $L_2$ norm of the attacks, we train the classifier with an adversary constrained to $\mathcal{S} = \{\delta: \norm{\delta}_2 < \epsilon \}$, for a given maximum perturbation $\epsilon$.

Lastly, we also consider a countermeasure using background removal. Handwritten data has an important difference compared to other vision tasks, such as object recognition, where we have a clear and simple separation of background and foreground. This is an important distinction because adversarial samples usually involve adding a crafted ``noise" all around the image. To this end, we investigate the impact (on the attack success rate) of removing the background after the adversarial samples are generated.

\section{Attack scenarios for Offline Handwritten Signature Verification}
\label{sec:attack_scenarios}

We now consider the possible attacks to biometric systems based on adversarial examples $\tilde{X}$. In particular, we identify possible attack points, and provide specific scenarios for Offline Handwritten Signature Verification.

The attacks using adversarial examples involve changing the inputs to the classifier, and therefore we identify two potential areas of vulnerability: at the sensor level, or the template storage/update level. The most prominent aspect of adversarial examples is that they fool a machine learning system without fooling humans (i.e. $\tilde{X}$ being visually similar to $X$). This is an important difference to spoofing attacks (that also target the sensor level), since these fake biometric traits, such as a ``gummy finger'', are clearly identified as different from a real finger by a human. We identify the following new attacks on signature verification systems, along with possible goals of an attacker:

\begin{enumerate}
	\item \textbf{Attacks on the data capture} (targets point \#1). In this case the adversarial image is crafted before the image is collected for the system. That is, an adversary can craft adversarial images $\tilde{X}$, and present them to the system, for instance using a banking application that allows a customer to use a picture of a cheque to cash it; or by printing adversarial noise on a physical signature. We identify two types of attack:
	\begin{itemize}
		\item \textbf{Type-I attack} (false rejection): Present a genuine signature that fools the system as being a forgery. This can be used for denial of service (preventing genuine users to accessing a system). We can also make a parallel to disguised signatures, where the user signs a document with the intent of later denying it (for example, the receiver of a check accepts it, but fails to cash it as the system classifies it is a forgery).
		\item \textbf{Type-II attack} (false acceptance): Present a random forgery (i.e. a genuine signature from user $y_i$) that fools the system as being genuine for user $y_j$ ($j \neq i$). At the same time, to a person, this sample can show no signs of being forgery (if it is not compared to a reference), since it is a genuine signature. The attacker can also use a skilled forgery as ``starting point", creating noise to increase the likelihood of the forgery being accepted. 
	\end{itemize}
	\item \textbf{Attacks on the templates} (targets point \#5): If original images are stored as part of the system (e.g. for classifier re-training, or manual verification in case of system failure/rejection of a sample), the templates can be changed to still look like genuine signatures for human operators, but in a way that accept signatures from a different person as genuine.
	\item \textbf{Attack on template update} (targets point \#7): For adaptive systems, the attacker can craft changes on the user's signature, so that adversarial templates are added to the gallery, to enable an intrusion later using a signature from another person. Similarly to the point above, the templates would appear as genuine to a person.
\end{enumerate}

The attacks above require different capabilities from the part of the attacker. The first attack only affects the system at test time (evasion attack), and in many practical scenarios would require the creation of a \emph{physical} attack, that is, the creation of an adversary signature in a piece of paper, for instance by printing adversarial noise on top of a handwritten signature. 
The second attack is a poisoning attack, that does not require a physical sample, as it impacts the stored templates of an user. However, it requires the capability of the attacker to update the template database, and can be categorized as an \emph{insider attack} as per the terminology used by Biggio et al \cite{biggio_adversarial_2015}. Note that this attack differ from simply adding another user's biometric to the templates, since a manual inspection of the templates would not reveal that the templates have been tampered with.
The third attack can also be seen as a poisoning attack, affecting adaptive systems, that automatically add new samples to the set of user templates. 

As for the knowledge required from the adversary, we can consider different scenarios, ranging from full knowledge of the system, to scenarios where only limited information is available to the attacker.

\subsection{Refining the adversary's knowledge model}
\label{sec:refining_knowledge}

For biometric \emph{verification} tasks, we identify an important refinement of the adversary's knowledge model. We argued in section \ref{sec:security_biometric} that an adversary that does not have access to the training set can collect its own data $\mathcal{\hat{D}}$ from the same data distribution, and train a surrogate classifier. For verification systems, each new user to the system effectively introduces a new class, and therefore it is important to make a distinction of accessing data for a particular individual of interest, and a ``background class", that are negative examples for a given user (e.g. signatures from other users). We refer therefore to two data components: $\mathcal{D}_b$ - biometric data from the background class (i.e. not for the individual under attack), and $\mathcal{D}_u$ - biometric data from the targeted individual. This allows the definition of limited knowledge scenarios where the biometric sample of the user can be collected, or for scenarios where the adversary can only collect samples from a other users. 

In our experiments, we consider three attack scenarios:

\begin{itemize}
	\item \textbf{Perfect Knowledge} scenario: the attacker has knowledge of all components: $\theta_{PK} = (\mathcal{D}_b, \mathcal{D}_u, \mathcal{X}, f, w)$. This scenario serves as a tool to analyze the worst-case scenario (from the system's defense perspective).
	\item \textbf{Limited Knowledge \#1}: we consider a scenario where the attacker does not have access to the dataset used for training the classifiers, but has access to all other components. We consider that the attacker is able to collect signatures from some users ($\mathcal{\hat{D}}_b$, that are from different users from those used to train the system), and some signatures from the user of interest, that were not used for training the system: $\mathcal{\hat{D}}_u$. In this case, $\theta_{LK1} = (\mathcal{\hat{D}}_b, \mathcal{\hat{D}}_u, \mathcal{X}, f, \hat{w})$.
	\item \textbf{Limited Knowledge \#2}: similarly to the above, but we consider a scenario where the attacker does not have full access to the feature extraction function (that induces the space $\mathcal{X}$). In particular, we consider a scenario where the attacker does not have access to the CNN model that was used to extract the features, but trains its own CNN (with identical training procedure and architecture) on a different set of users. In this case, $\theta_{LK2} = (\mathcal{\hat{D}}_b, \mathcal{\hat{D}}_u, \mathcal{\hat{X}}, f, \hat{w})$.
\end{itemize}

\section{Experimental Protocol}
\label{sec:experimental}

\begin{figure}
	\centering
	\includegraphics[width=\columnwidth]{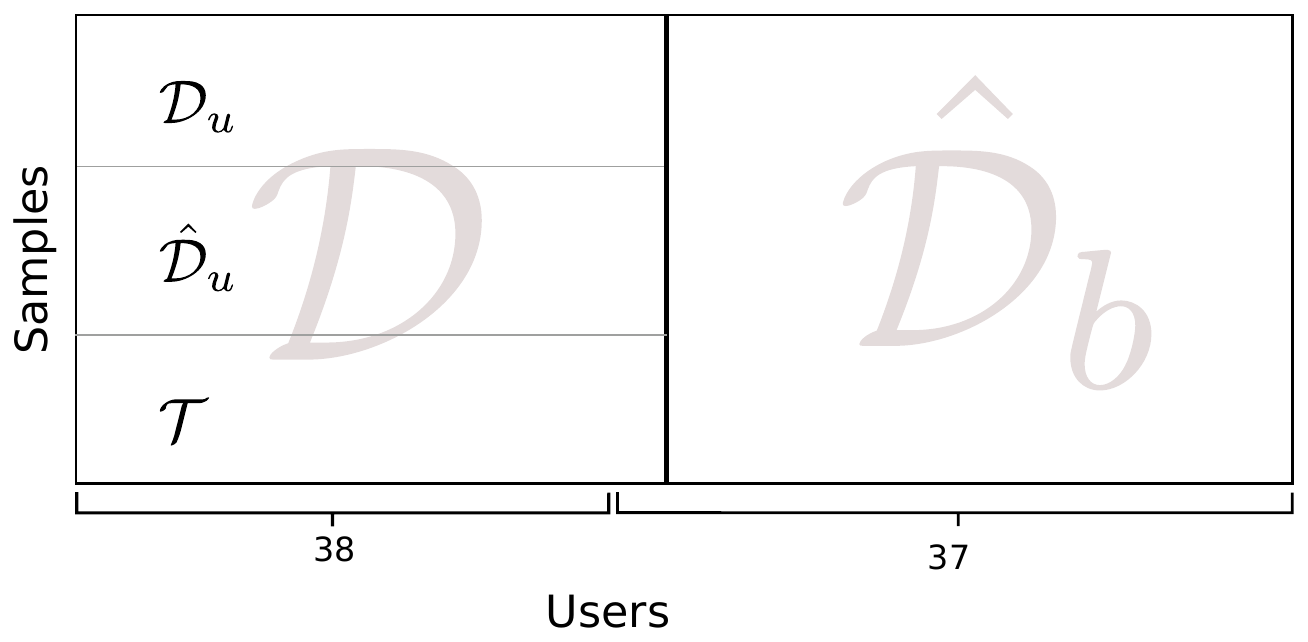}
	\caption{Dataset separation for the MCYT dataset. The set $\mathcal{D}_u$ is used for training the classifiers under attack, and the sets  $\mathcal{\hat{D}}_b$ and $\mathcal{\hat{D}}_u$ are used by the attacker to train surrogate classifiers.}
	\label{fig:dataset_separation}
\end{figure}

We conducted experiments using the datasets MCYT-75 \cite{ortega-garcia_mcyt_2003} (with 75 users), CEDAR \cite{kalera_offline_2004} (55 users), GPDS-160  \cite{vargas_off-line_2007} (160 users) and the Brazilian PUC-PR \cite{freitas_bases_2000} (60 users).

In order to simulate the different attack scenarios we split the dataset into two parts of disjoint users, as illustrated in Figure \ref{fig:dataset_separation}. The set $\mathcal{D}$ refers to the users ``enrolled in the system", that will be under attack. This dataset is divided in training (user signatures $\mathcal{D}_u$) and testing $\mathcal{T}$. For the limited knowledge scenarios, we consider a set $\mathcal{\hat{D}}_b$ that contains signatures from other users (not those being attacked), simulating the scenario of an attacker that acquired his own signature dataset in order to generate the attacks. We also consider that the attacker has access to some signatures from the user, $\mathcal{\hat{D}}_u$, that were not used for training the system (i.e. disjoint from $\mathcal{D}_u$ and $\mathcal{T}$).

The images were pre-processed in a similar way to \cite{hafemann_learning_2017}: Signatures were first centered in a blank canvas using their center of mass. We then resize the images to $150 \times 220$ pixels and invert the image such that the background pixels are zero-valued. Lastly, we run the OTSU algorithm to identify the optimal threshold that separates background and foreground. We set the pixels with intensity smaller than the threshold to intensity 0, leaving the remaining pixels in grayscale.

We consider Writer-Dependent classifiers, training an SVM (linear or with the RBF kernel) for each user. As feature extraction $\phi(X)$, we consider: (i) a CNN-based learned representation: SigNet \cite{hafemann_learning_2017}, and (ii) the CLBP operator (Completed Local Binary Patterns)\cite{guo_completed_2010}. We train the SVMs with 5 genuine signatures from the user as positive samples, and 5 signatures from each other user as negative. 

For the scenario LK2, we consider two CNN models with the same architecture and training procedures, but trained on a disjoint set of users. The CNN used by the model under attack was trained on GPDS users 350-614, and the CNN used by the surrogate models (by the attacker) were trained with users 615-881. Training procedure followed the same as SigNet \cite{hafemann_learning_2017}. For the Ensemble Adversarial Learning evaluation, we first trained two models with different architectures (slight variations from SigNet, as described in the Supplemental Material) and then trained a model with the SigNet architecture and the loss function defined in equation \ref{eq:ensadv}, with $\epsilon = 5$. For the Madry defense, we also used the same architecture, and trained with $\mathcal{S} = \{\delta : \norm{\delta}_2 < \epsilon\}$ with $\epsilon = 2$. We tried using larger values for $\epsilon$ and obtained worse classification performance during the CNN training, so these values represent a tradeoff between robustness and accuracy. In both cases, we trained the network with users 350-614, to enable evaluating the scenario LK2. In this scenario, we consider an attacker that trained a regular CNN (no adversarial training), with users 615-881.

After training the classifiers for each user, the SVMs implement the following decision functions:

\begin{align}
s_\text{Linear} &= \textbf{w}^\intercal \phi(X) + b \\
s_\text{RBF} &= \sum_{i \in \mathcal{S}}{\alpha_i \exp(-\gamma \norm{\phi(X) - X_i}) + b}
\end{align}

\noindent Where $s_\text{Linear}$ and $s_\text{RBF}$ are the scores for the linear SVM and the SVM with the RBF kernel, respectively; $w$ are the weights learned by the linear SVM, $\mathcal{S}$ is the set of support vectors, $\alpha_i$ and $X_i$ are the coefficients and support vectors, $\gamma$ is a hyperparameter for the RBF kernel and $b$ is the bias. We can easily see that both functions are differentiable with respect to  $\phi(X)$ \cite{biggio_evasion_2013}. For the classifier using a CNN-based model to extract the features, we can calculate the gradients of the scores w.r.t the inputs X, and apply gradient-based methods to generate the attacks. For non-differentiable feature extractors, we consider only the two gradient-free methods described in section \ref{sec:attacks_considered}. When reporting the scores in Figures \ref{fig:adv_example}, \ref{fig:cnn_attacks} and \ref{fig:lbp_attacks}, we consider a normalized loss as follows: $\tilde{s}(X) = s(X) - \tau$, where $\tau$ is the global threshold. This makes it easy to identify if a signature would be classified as genuine or as a forgery ($\tilde{s}(X) \ge 0$ indicates the prediction of $X$ being a genuine signature).

For the classifiers using LBP, we consider the the operator CLBP\_S/M/C \cite{guo_completed_2010} (3D histogram of CLBP S, M and C), with the following parameters: $R=1$ (radius of 1 pixel), $P=8$ (eight neighbors) and rotation invariant uniform patterns (``riu2"). The feature vector has a total of 200 dimensions.

To simplify the generation of the attacks we considered a global threshold for the classifications, that obtained the Equal Error Rate on the set $\mathcal{D}$ (without any attacks).

After the classifiers are trained, we generate attacks using the four methods described in section \ref{sec:attacks_considered}. We used the FGM method with $\epsilon = 1000$, and the Carlini \& Wagner attack with $\kappa = 1$. For the Decision-based attack, we considered the implementation from the authors\footnote{\url{https://github.com/bethgelab/foolbox}}, running the attack for a maximum of 1000 iterations. For the Simulated Annealing method, we considered an open implementation of simulated annealing\footnote{\url{https://github.com/perrygeo/simanneal}}. In each iteration, we change the state by adding gaussian noise $\epsilon$ ($\epsilon \sim \mathcal{N}(0, \sigma I)$, with $\sigma = 2$), and clipping the image between 0 and 255. We consider the energy to be a mixture of the SVM score and the $L_2$ norm of the adversarial noise $\delta$: $E = s(X) + \lambda \norm{\delta}_2$, with $\lambda = 0.001$ being a trade-off between changing the SVM score, and not deviating too far from the original image. We used an initial temperature $T_\text{max}=1$ and final temperature $T_\text{min} = 0.001$. These values were chosen such that around 95\% of the steps that would increase the energy are still accepted in the start of the procedure, and less than 5\% were accepted in the end. We ran this procedure with at most 1000 steps, with early stopping (we stop optimization if the image is adversarial).

The experiments consisted in Type-I attacks (attempting to have a genuine signature rejected by the system) and Type-II attacks (attempting to have a forgery accepted by the system). For each user, we selected one genuine signature, one random forgery and one skilled forgery, such that all four classifiers correctly classified these samples. We then used the different attack methods to generate adversarial samples, and measured the attack success rate (number of misclassified images after the attack), and the average RMSE (root mean square error) of the adversarial noise on successful attacks. It is worth noting that we consider pixel values in the range $[0, 255]$, so the RMSE of the adversarial noise is also constrained in the same range. To summarize the experiments, we considered:

\begin{itemize}
	\item Datasets: MCYT-75, CEDAR, GPDS-160, Brazilian PUC-PR
	\item Feature extractor: CLBP, SigNet
	\item SVM type: Linear, RBF
	\item Attack method: Decision-based, Simulated Annealing, FGM, Carlini
	\item Attacker's goal: Type-I (attacking Genuine signatures) and Type-II (attacking Random and Skilled forgeries), 
	\item Attacker's knowledge: Perfect Knowledge, Limited Knowledge LK1 and LK2
	\item Defense: No defense, Ens. Adv. training, Madry
\end{itemize}

It is worth mentioning that in this work we did not consider the discretization of the generated adversarial images. We worked with images in float format, instead of discretized into integers between 0 and 255. This is discussed in section \ref{sec:practical}.

\section{Results and discussion}
\label{sec:results}

\begin{table}
	\centering
	\caption{Results of WD classifiers using different feature sets (EER considering skilled forgeries)}
	\label{tbl:perf}
	\begin{tabular}{llrrrr}
		\toprule
		Dataset & Features &  \multicolumn{2}{c}{EER global-$\tau$} & \multicolumn{2}{c}{EER user-$\tau$}  \\
		& & Linear & RBF & Linear & RBF \\
		\midrule
		MCYT-75 &     SigNet &  7.12 &      7.03 &  7.39 &     5.68\\
		 &        CLBP &  26.49 &      27.03 &  27.21 &    26.85\\
		CEDAR &     SigNet & 12.03 & 11.82 &  6.01 &   4.52\\
		 &        CLBP & 28.01 & 21.36 &  23.95 & 16.39\\
		GPDS & Signet & 7.70 &  6.80 &  4.62 &   4.14\\
		 & CLBP & 26.74 & 24.58 &  21.79 & 22.37\\
		Brazilian PUC-PR &  SigNet &  6.78 &   5.22 &  3.61 &    2.67\\
		&        CLBP &  26.83 &     19.61 &  24.61 &      16.83\\
		
		\bottomrule
	\end{tabular}
\end{table}

Before presenting the results of the attacks, we first validate the performance of the WD classifiers on the four datasets. Table \ref{tbl:perf} shows the EER obtained by using different features/classifiers, when trained with 5 reference signatures per user, with the protocol defined in section \ref{sec:experimental}. We observe a large variance in the results across different datasets, which suggests different degrees of difficulty on separating genuine signatures and forgeries in them. We also observe a large difference of performance between systems trained with the SigNet and CLBP features. In order to have a fair analysis of the adversarial examples against each classifier/feature extractor, we select the same set of images for the attacks on all classifiers, ensuring that the original images (before the attack) were correctly classified by them. Although the classifier performance varies across different datasets, the results for the adversarial attacks showed consistent trends across them. In this paper we report the consolidated results over the four datasets, and for completeness we include the results on individual datasets in the Supplementary Material.

\subsection{Perfect Knowledge}
\label{sec:result_pk}

We consider first a scenario of Perfect Knowledge, in which the adversary has full knowledge of all components of the system: $\theta_{PK} = (\mathcal{D}_b, \mathcal{D}_u, \mathcal{X}, f, w)$. The attacker can run his own copy of the system, and use one of the proposed attacks to generate adversarial images. 

\begin{table}
	\centering
	\caption{Success rate of Type-I attacks (\% of attacks that transformed a genuine signature in a forgery)}
	\label{tbl:success_evasion}
	\begin{tabular}{ll|rrrr}
		\toprule
		& & \multicolumn{4}{c}{Attack Type} \\
		Feature &  Classifier &  FGM & Carlini & Anneal & Decision \\
		\midrule
		CLBP & Linear   &     - &        - &   63.16 &     80.70 \\
		CLBP & RBF      &     - &        - &  100.00 &    100.00 \\
		SigNet & Linear & 99.42 &   100.00 &   98.83 &    100.00 \\
		SigNet & RBF    & 98.25 &   100.00 &   98.83 &    100.00 \\
		
		\bottomrule
	\end{tabular}
\end{table}

\begin{table}
	\centering
	\caption{Distortion (RMSE of the adversarial noise) for successful Type-I attacks}
	\label{tbl:mse_evasion}
	\begin{tabular}{ll|rrrr}
		\toprule
		& & \multicolumn{4}{c}{Attack Type} \\
		Feature &  Classifier &  FGM & Carlini & Anneal & Decision \\
		\midrule
		CLBP & Linear   &    - &        - &    0.40 &      1.57 \\
		CLBP & RBF      &    - &        - &    0.36 &      $10^{-9}$ \\
		SigNet & Linear & 4.04 &     1.35 &    5.69 &      3.27 \\
		SigNet & RBF    & 4.06 &     1.40 &    5.17 &      3.02 \\
		\bottomrule
	\end{tabular}
\end{table}

\begin{figure}
	\centering
	\subfloat[Carlini ($\tilde{s} = -0.69$, RMSE 2.46)]{
		\includegraphics[scale=0.5]{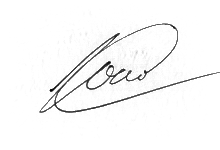}}
	\qquad
	\subfloat[FGM ($\tilde{s} = -0.79$, RMSE 3.48)]{
		\includegraphics[scale=0.5]{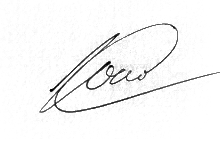}}
	\qquad
		\subfloat[Decision ($\tilde{s} = -0.65$, RMSE 6.21)]{
			\includegraphics[scale=0.5]{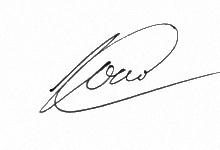}}
		\quad
		\subfloat[Anneal ($\tilde{s} = -0.62$, RMSE 7.66)]{
			\includegraphics[scale=0.5]{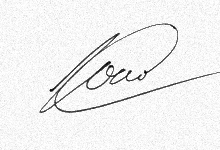}}
	\caption{Example of Type-I attacks on the SVM model with RBF kernel and SigNet features. The original image is correctly classified as genuine by this model ($\tilde{s} = 0.13$).}
	\label{fig:cnn_attacks}
\end{figure}

\begin{figure}
	\centering
	\subfloat[Decision ($\tilde{s} = -0.39$, RMSE 0.99)]{
		\includegraphics[scale=0.5]{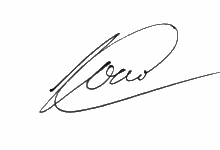}}
	\qquad
	\subfloat[Anneal ($\tilde{s} = -0.27$, RMSE 0.21)]{
		\includegraphics[scale=0.5]{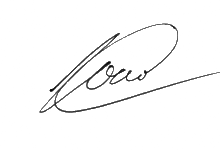}}
	\caption{Example of Type-I attacks on the SVM model with Linear kernel and CLBP features. The original image is correctly classified as genuine by this model ($\tilde{s} = 1.60$).}		\label{fig:lbp_attacks}
\end{figure}

For Type-I attacks, given a genuine sample $X_g$, the objective is to obtain an adversarial $\tilde{X} = X_g + \delta$ that is classified as a forgery.  Table \ref{tbl:success_evasion} shows the success rate of attacks in this scenario (i.e. the percentage of attacks that found an adversary image), by attack type and classifier type. We see a high success rate for most attacks. Table \ref{tbl:mse_evasion} shows the average RMSE (root mean squared error) of the adversarial noise $\delta$. We notice that the required amount of noise varies significantly with different classifiers and attack types. In general, gradient-based attacks find adversarial images with much less noise on the differentiable models. For the models with handcrafted features (where we do not have gradients), we noticed that even smaller changes on the image were enough to induce a misclassification. Figures \ref{fig:cnn_attacks} and \ref{fig:lbp_attacks} present examples of this type of attack.

\begin{table}
	\centering
	\caption{Success rate of Type-II attacks (\% of attacks that transformed a forgery in a genuine signature)}
	\label{tbl:success_intrusion}
	\resizebox{\columnwidth}{!}{%
		\begin{tabular}{lll|rrrr}
			\toprule
			&&& \multicolumn{4}{c}{Attack Type} \\
			&  &&   FGM &  Carlini &  Anneal &  Decision \\
			Features & Classifier & Forgery Type &       &          &         &           \\
			\midrule
			CLBP & Linear & random &     - &        - &   37.36 &     45.98 \\
			&& skilled &     - &        - &   38.73 &     46.24 \\
			CLBP & RBF & random &     - &        - &    0.00 &      0.00 \\
			&& skilled &     - &        - &    0.00 &      0.00 \\
			SigNet & Linear & random &  1.15 &    96.55 &    0.00 &      0.00 \\
			&& skilled & 28.90 &    99.42 &    2.31 &      3.47 \\
			SigNet & RBF & random &  0.57 &    94.83 &    0.00 &      0.00 \\
			&& skilled & 19.65 &   100.00 &    1.73 &      1.73 \\
			
			\bottomrule
		\end{tabular}
	}
	
\end{table}

\begin{table}
	\centering
	\caption{Distortion (RMSE of the adversarial noise) for successful Type-II attacks}
	\label{tbl:mse_intrusion}
	\resizebox{\columnwidth}{!}{%
		\begin{tabular}{lll|rrrr}
			\toprule
			&&& \multicolumn{4}{c}{Attack Type} \\
			&  &&   FGM &  Carlini &  Anneal &  Decision \\
			Features & Classifier & Forgery Type &       &          &         &           \\
			\midrule
			CLBP & Linear & random &    - &        - &    0.39 &      1.17 \\
			&& skilled &    - &        - &    0.42 &      1.08 \\
			SigNet & Linear & random & 4.11 &     6.07 &       - &         - \\
			&& skilled & 4.20 &     3.19 &    3.61 &      1.34 \\
			SigNet & RBF & random & 4.70 &     6.55 &       - &         - \\
			&& skilled & 4.08 &     3.62 &    3.17 &      1.18 \\
			
			\bottomrule
		\end{tabular}
	}
\end{table}

We now consider Type-II attacks, in which we want to modify a forgery $X_f$, by creating an adversary $\tilde{X} = X_f + \delta$ that is classified as a genuine signature. Table \ref{tbl:success_intrusion} shows the success rate of the different methods, and table \ref{tbl:mse_intrusion} shows the level of noise required in the successful attacks. The results show that this attack is much harder to obtain compared to the Type-I attacks. For the models trained with CLBP features, we observed that the linear classifier could be attacked half of the time, while we could not generate any attack using the two gradient-free methods for the non-linear model. For the CNN-based models, a strong gradient-based method (Carlini) worked for almost all samples, while the gradient-free methods did not work in most cases - we only observed some success when using skilled forgeries as the starting point. Comparing tables \ref{tbl:mse_evasion} and \ref{tbl:mse_intrusion}, we observe that for the CLBP-based classifiers, a similar amount of noise was required to create successful attacks. For the CNN-based methods, when starting from a random forgery a large amount of noise was required to create successful attacks, while when starting from a skilled forgery a lower amount of noise was required. We reiterate that the skilled forgeries selected for attack were correctly classified by the model (i.e. classified as forgeries), while in successful attacks the adversarial image is classified as a genuine.

It is worth noting that in the experiments with the strong gradient-based attack, we observed a much larger amount of noise required for misclassification compared to previous results reports on object recognition. For instance, in the classification task on ImageNet, successful attacks (using the same Carlini \& Wagner method) are reported with much lower noise (RMSE of 0.004 for 100\% success of targeted attacks on ImageNet \cite{carlini_towards_2017}). While for object recognition the adversarial images are often perceptually identical to the original, for signatures we noted some distinguishable noise, specially on the Type-II attacks, as can be seen in figure \ref{fig:adv_example} (where the Type-II attack has RMSE of 10.34).

\subsection{Limited Knowledge \#1}
\label{sec:result_lk1}

We now consider a limited knowledge scenario, where the attacker does not have access to the signatures used for training the system, but does obtain a surrogate dataset: $\theta_{LK1} = (\mathcal{\hat{D}}_b, \mathcal{\hat{D}}_u, \mathcal{X}, f, \hat{w})$. In this case, the signatures from the \emph{background set} (used as negative samples during training) were from a different set of users than those used to train the system. We also consider that the attacker collected some signatures from the user of interest $\mathcal{\hat{D}}_u$, but that are also different from those used to train the system. This scenario also assumes that the attacker knows the feature extractor (i.e. full knowledge of the feature extractor, including all parameters), and the learning function (the WD classifier type, but not the learned parameters). In this scenario, the attacker uses the surrogate data to train their own version of the WD classifiers, and uses this classifier to generate the attacks. We then evaluate the success rate of these attacks on the actual system.

\begin{table}
	\centering
	\caption{Success rate of Type-I attacks (\% of attacks that transformed a genuine signature in a forgery)  (Limited Knowledge)}
	\label{tbl:success_evasion_lk1}
	\begin{tabular}{ll|rrrr}
		\toprule
		& & \multicolumn{4}{c}{Attack Type} \\
		Feature &  Classifier &  FGM & Carlini & Anneal & Decision \\
		\midrule
		CLBP & Linear   &     - &        - &   42.69 &     43.86 \\
		CLBP & RBF      &     - &        - &   82.46 &     82.46 \\
		SigNet & Linear & 97.08 &    80.12 &   50.88 &     40.35 \\
		SigNet & RBF    & 97.66 &    91.81 &   54.39 &     47.95 \\
		\bottomrule
	\end{tabular}
\end{table}

Table \ref{tbl:success_evasion_lk1} shows the success rate of the Type-I attacks. We observe a lower success rate compared to the perfect knowledge scenario, but still we find a high success rate against most models. This suggests that indeed there is a transferability of attacks across models (as observed before in CNNs \cite{szegedy_intriguing_2013}), and that this transferability also impacts systems trained with handcrafted features.

\begin{table}
	\centering
	\caption{Success rate of Type-II attacks (\% of attacks that transformed a forgery in a genuine signature)  (Limited Knowledge)}
	\label{tbl:success_intrusion_lk1}
	\resizebox{\columnwidth}{!}{%
		\begin{tabular}{lll|rrrr}
			\toprule
			&&& \multicolumn{4}{c}{Attack Type} \\
			&  &&   FGM &  Carlini &  Anneal &  Decision \\
			Features & Classifier & Forgery Type &       &          &         &           \\
			\midrule
			
			CLBP & Linear & random &     - &        - &   24.71 &     28.74 \\
			&& skilled &     - &        - &   21.97 &     26.01 \\
			CLBP & RBF & random &     - &        - &    0.00 &      0.00 \\
			&& skilled &     - &        - &    0.00 &      0.00 \\
			SigNet & Linear & random &  0.00 &    46.55 &    0.00 &      0.00 \\
			&& skilled & 22.54 &    71.68 &    1.73 &      0.00 \\
			SigNet & RBF & random &  0.00 &    74.71 &    0.00 &      0.00 \\
			&& skilled & 19.08 &    83.24 &    0.58 &      0.00 \\
			
			\bottomrule
		\end{tabular}
	}	
\end{table}

Table \ref{tbl:success_intrusion_lk1} shows the success rate for Type-II attacks in a limited knowledge scenario. Again we see a drop in performance compared to the perfect knowledge scenario, but still the attacks that worked in the PK scenario also worked (to some extent) in the limited knowledge scenario.

\subsection{Limited Knowledge \#2}

We now consider a limited knowledge scenario similar to the above, but where the attacker also does not have access to the CNN used to extract the features. In this case, we consider that the attacker trains a surrogate CNN using a disjoint set of users, which induces a new feature space $\mathcal{\hat{X}}$. We consider therefore  $\theta_{LK2} = (\mathcal{\hat{D}}_b, \mathcal{\hat{D}}_u, \mathcal{\hat{X}}, f, \hat{w})$. 

\begin{table}
	\centering
	\caption{Success rate of Type-I attacks (\% of attacks that transformed a genuine signature in a forgery)  (Limited Knowledge \#2)}
	\label{tbl:success_evasion_lk2}
	\begin{tabular}{ll|rrrr}
		\toprule
		& & \multicolumn{4}{c}{Attack Type} \\
		Feature &  Classifier &  FGM & Carlini & Anneal & Decision \\
		\midrule
			SigNet & Linear & 60.34 &     6.90 &   48.85 &     19.54 \\
			SigNet & RBF    & 64.37 &     9.20 &   51.15 &     18.97 \\
		\bottomrule
	\end{tabular}
\end{table}

\begin{table}
	\centering
	\caption{Success rate of Type-II attacks (\% of attacks that transformed a forgery in a genuine signature)  (Limited Knowledge \#2)}
	\label{tbl:success_intrusion_lk2}
	\resizebox{\columnwidth}{!}{%
		\begin{tabular}{lll|rrrr}
			\toprule
			&&& \multicolumn{4}{c}{Attack Type} \\
			&  &&   FGM &  Carlini &  Anneal &  Decision \\
			Features & Classifier & Forgery Type &       &          &         &           \\
			\midrule
			SigNet & Linear & random & 0.00 &     0.00 &    0.00 &      0.00 \\
			&& skilled & 2.30 &     2.30 &    0.57 &      0.57 \\
			SigNet & RBF & random & 0.00 &     0.00 &    0.00 &      0.00 \\
			&& skilled & 1.72 &     1.72 &    1.15 &      0.00 \\

			\bottomrule
		\end{tabular}
	}	
\end{table}

Tables \ref{tbl:success_evasion_lk2} \ref{tbl:success_intrusion_lk2} show the success rate of the Type-I and Type-II attacks, respectively. We observe much lower success rates, especially for Type-II attacks, where no attacks were successful when starting from a random forgery, and starting with a skilled forgery the success was as low as 1-2\%. For the Type-I attacks, we notice lower success rates compared to the PK and LK1 scenarios. Overall, these results suggest that transferability of the attacks is much worse when the models are trained with a different subset of users, that is, when the attacker does not have access to signatures from the same users that were used to train the CNN model. This contrasts with findings in object classification, where attacks trained on a subset of data  transfer well to a model trained with another subset of data (different samples from the same classes) \cite{szegedy_intriguing_2013}. Also, it is worth noting that the strong Carlini attack (that achieves close to 100\% success in the Perfect Knowledge scenario) drops in performance in the LK scenarios, confirming previous findings that such iterative attacks transfer less than single-step attacks such as FGM \cite{kurakin_adversarial_2016-1}.

\subsection{Evaluating countermeasures}

\begin{table*}
	\centering
	\caption{Success rate of Type-I attacks considering different defenses and attacker knowledge scenarios}
	\label{tbl:success_defenses_evasion}
	\begin{tabular}{ll|rrr|rrr}
		\toprule
			&& \multicolumn{6}{c}{Attack Type and Knowledge scenario} \\
&  &   \multicolumn{3}{c}{FGM} & \multicolumn{3}{c}{Carlini}\\
		Defense & Classifier & PK &   LK1 &   LK2 &     PK &   LK1 &  LK2 \\
		\midrule
		Baseline &  Linear & 100.00 & 95.40 & 60.34 &  100.00 & 78.16 & 6.90 \\
		& RBF & 100.00 & 97.70 & 64.37 &  100.00 & 85.63 & 9.20 \\
		Ens. Adv &  Linear &  91.38 & 85.63 & 45.40 &  100.00 & 79.89 & 4.60 \\
		&  RBF &  90.23 & 83.91 & 46.55 &  100.00 & 90.23 & 5.75 \\
		Madry &  Linear &  91.38 & 83.33 & 22.99 &  100.00 & 74.71 & 1.72 \\
		& RBF &  89.08 & 86.21 & 21.84 &  100.00 & 89.08 & 0.57 \\
		\bottomrule
	\end{tabular}
\end{table*}

\begin{table*}
	\centering
	\caption{Distortion (RMSE of the adversarial noise) for Type-I attacks, considering different defenses and attacker knowledge scenarios}
	\label{tbl:mse_defenses_evasion}
	\begin{tabular}{ll|rrr|rrr}
	\toprule
	&& \multicolumn{6}{c}{Attack Type and Knowledge scenario} \\
	&  &   \multicolumn{3}{c}{FGM} & \multicolumn{3}{c}{Carlini}\\
	Defense & Classifier & PK &   LK1 &   LK2 &     PK &   LK1 &  LK2 \\
			\midrule
			Baseline &  Linear & 4.17 & 4.19 & 4.30 &    1.31 & 1.33 & 1.37 \\
			&  RBF & 4.20 & 4.21 & 4.30 &    1.40 & 1.38 & 1.55 \\
			Ens. Adv. & SigNet \& Linear & 4.37 & 4.30 & 4.20 &    1.35 & 1.43 & 1.85 \\
			&  RBF & 4.36 & 4.32 & 4.20 &    1.44 & 1.43 & 1.63 \\
			Madry & SigNet \& Linear & 4.76 & 4.72 & 4.26 &    3.19 & 3.28 & 1.59 \\
			&  RBF & 4.77 & 4.74 & 4.27 &    3.48 & 3.52 & 2.19 \\
			\bottomrule
		\end{tabular}
\end{table*}

We now consider the impact of two counter-measures for the CNN-based systems: Ensemble Adversarial Learning \cite{tramer_ensemble_2017} and the Madry defense \cite{madry_towards_2017}. Tables \ref{tbl:success_defenses_evasion} and \ref{tbl:mse_defenses_evasion} show the success rate and distortion (RMSE) for Type-I attacks. We consider the three Knowledge scenarios discussed in section \ref{sec:refining_knowledge} (Perfect Knowledge and two Limited Knowledge scenarios), and the two gradient-based attacks (FGM and Carlini). We notice that both defenses provide some robustness against the FGM attack in all knowledge scenarios. Considering the Carlini attack, we see that in a Perfect-Knowledge scenario the attack was always successful, but Table \ref{tbl:mse_defenses_evasion} shows that the Madry defense greatly increase the amount of noise required to generate adversarial examples, going from a RMSE of 1.4 to around 3.3.

\begin{table*}
	\centering
	\caption{Success rate of Type-II attacks considering different defenses and attacker knowledge scenarios}
	\label{tbl:success_defenses_intrusion}
	
	\begin{tabular}{lll|rrr|rrr}
		\toprule
&&& \multicolumn{6}{c}{Attack Type and Knowledge scenario} \\
&&  &   \multicolumn{3}{c}{FGM} & \multicolumn{3}{c}{Carlini}\\
Defense & Classifier & Forgery Type & PK &   LK1 &   LK2 &     PK &   LK1 &  LK2 \\
		\midrule
		Baseline & Linear & random &  2.87 &  1.15 & 0.00 &   98.85 & 42.53 & 0.00 \\
		&              & skilled & 40.80 & 29.31 & 2.30 &  100.00 & 66.67 & 2.30 \\
		& RBF & random &  1.72 &  1.15 & 0.00 &   95.98 & 68.39 & 0.00 \\
		&              & skilled & 34.48 & 27.59 & 1.72 &  100.00 & 83.91 & 1.72 \\
		Ens. Adv. & Linear & random &  1.72 &  0.57 & 0.00 &   93.10 & 41.38 & 0.00 \\
		&              & skilled & 29.31 & 14.94 & 1.15 &  100.00 & 64.37 & 3.45 \\
		& RBF &  random &  2.30 &  0.00 & 0.00 &   93.10 & 69.54 & 0.00 \\
		&              & skilled & 22.99 & 17.24 & 1.15 &  100.00 & 83.91 & 2.30 \\
		Madry &  Linear & random &  1.72 &  0.57 & 0.00 &   98.28 & 45.98 & 0.00 \\
		&              & skilled & 48.85 & 38.51 & 8.05 &  100.00 & 73.56 & 3.45 \\
		&  RBF & random &  2.30 &  0.57 & 0.00 &   97.70 & 75.86 & 0.00 \\
		&              & skilled & 45.98 & 37.36 & 6.32 &  100.00 & 87.93 & 2.87 \\
		\bottomrule
	\end{tabular}
\end{table*}

\begin{table*}
	\centering
	\caption{Distortion (RMSE of the adversarial noise) for Type-II attacks, considering different defenses and attacker knowledge scenarios}
	\label{tbl:mse_defenses_intrusion}
	\begin{tabular}{lll|rrr|rrr}
	\toprule
	&&& \multicolumn{6}{c}{Attack Type and Knowledge scenario} \\
	&&  &   \multicolumn{3}{c}{FGM} & \multicolumn{3}{c}{Carlini}\\
	Defense & Classifier & Forgery Type & PK &   LK1 &   LK2 &     PK &   LK1 &  LK2 \\
		\midrule
		Baseline & Linear & random & 3.97 & 3.75 &    - &    5.98 &  5.71 &    - \\
		&              & skilled & 4.21 & 4.14 & 4.24 &    2.99 &  2.71 & 2.43 \\
		&  RBF & random & 3.84 & 3.83 &    - &    6.27 &  6.03 &    - \\
		&              & skilled & 4.11 & 4.06 & 4.64 &    3.32 &  3.20 & 1.77 \\
		Ens. Adv. &  Linear & random & 4.51 & 4.82 &    - &    8.61 &  8.83 &    - \\
		&              & skilled & 4.53 & 4.58 & 4.09 &    4.71 &  4.34 & 1.43 \\
		&  RBF & random & 4.40 &    - &    - &    9.45 &  9.31 &    - \\
		&              & skilled & 4.59 & 4.58 & 4.07 &    5.43 &  4.82 & 2.14 \\
		Madry &  Linear & random & 4.74 & 5.38 &    - &   10.81 & 10.97 &    - \\
		&              & skilled & 4.90 & 4.93 & 4.15 &    6.18 &  5.87 & 1.94 \\
		&  RBF & random & 4.62 & 5.28 &    - &   11.49 & 11.46 &    - \\
		&              & skilled & 4.91 & 4.88 & 4.16 &    7.00 &  6.71 & 2.40 \\
		\bottomrule
	\end{tabular}
\end{table*}

Tables \ref{tbl:success_defenses_intrusion} and \ref{tbl:mse_defenses_intrusion} shows the results on Type-II attacks. In these experiments, we again observe that the Carlini attack finds attacks most of the time, and that the Madry defense showed to be effective in increasing the amount of noise required to obtain an adversarial example (e.g. the average RMSE is increased from 5.98 to 10.81 when starting with a random forgery, comparing the baseline and the Madry defense). It is worth noting that the RMSE values only consider the successful attacks, and therefore the results on the Limited Knowledge scenarios (where the success rate is very low) are likely skewed by a few forgeries that were already close to the decision boundary.

\subsection{Impact of background removal}
\label{sec:noise_removal}

\begin{table*}
	\centering
\caption{Success of Type-I attacks in a PK scenario, with no pre-processing and with OTSU pre-processing}
\label{tbl:otsu_evasion}

	\begin{tabular}{ll|rrrrrrrr}
	\toprule
	&& \multicolumn{8}{c}{Attack Type and Preprocessing} \\
	& & \multicolumn{2}{c}{FGM} & \multicolumn{2}{c}{Carlini} & \multicolumn{2}{l}{Anneal} & \multicolumn{2}{c}{Decision} \\
	&  & None & OTSU & None & OTSU & None & OTSU & None & OTSU\\
	Feature & Classifier  &          &              &         &              &         &              &          &              \\
		\midrule
		CLBP & Linear &       - &            - &       - &            - &   63.16 &         9.36 &    80.70 &         3.51 \\
		& RBF &       - &            - &       - &            - &  100.00 &         0.58 &   100.00 &         0.00 \\
		SigNet baseline &  Linear &  100.00 &        88.51 &  100.00 &        18.39 &   96.55 &         0.57 &   100.00 &         1.72 \\
		&  RBF &  100.00 &        86.21 &  100.00 &        22.41 &   98.28 &         0.00 &    98.85 &         0.57 \\
		SigNet Ens. Adv. &  Linear &   91.38 &        67.24 &  100.00 &         2.87 &   97.70 &         0.00 &   100.00 &         0.00 \\
		& RBF &   90.23 &        65.52 &  100.00 &         1.72 &   98.28 &         0.00 &   100.00 &         0.00 \\
		SigNet Madry & Linear &   91.38 &        87.93 &  100.00 &        77.01 &   87.93 &         0.00 &    99.43 &         6.90 \\
		& RBF &   89.08 &        87.36 &  100.00 &        75.86 &   88.51 &         0.00 &   100.00 &         5.75 \\
		\bottomrule
	\end{tabular}
\end{table*}

\begin{table*}
		\centering
	\caption{Success of Type-II attacks in a PK scenario, with no pre-processing and with OTSU pre-processing}
	\label{tbl:otsu_intrusion}
	\begin{tabular}{lll|rrrrrrrr}
	\toprule
	&&& \multicolumn{8}{c}{Attack Type and Preprocessing} \\
	&& & \multicolumn{2}{c}{FGM} & \multicolumn{2}{c}{Carlini} & \multicolumn{2}{l}{Anneal} & \multicolumn{2}{c}{Decision} \\
	&&  & None & OTSU & None & OTSU & None & OTSU & None & OTSU\\
	Feature & Classifier  & Forgery Type &         &              &         &              &         &              &          &              \\
		\midrule
		CLBP & Linear & random &       - &            - &       - &            - &   37.36 &         0.57 &    45.98 &         0.00 \\
		&              & skilled &       - &            - &       - &            - &   38.73 &         1.73 &    46.24 &         1.16 \\
		& RBF & random &       - &            - &       - &            - &    0.00 &         0.00 &     0.00 &         0.00 \\
		&              & skilled &       - &            - &       - &            - &    0.00 &         0.00 &     0.00 &         0.00 \\
		SigNet Baseline &  Linear & random &    2.87 &         0.57 &   98.85 &         0.00 &    0.00 &         0.00 &     0.00 &         0.00 \\
		&              & skilled &   40.80 &        31.03 &  100.00 &        12.07 &    1.15 &         0.00 &     1.15 &         0.00 \\
		&  RBF & random &    1.72 &         0.57 &   95.98 &         0.00 &    0.00 &         0.00 &     0.00 &         0.00 \\
		&              & skilled &   34.48 &        24.14 &  100.00 &        14.37 &    1.72 &         0.00 &     1.72 &         0.00 \\
		SigNet Ens Adv. &  Linear & random &    1.72 &         1.15 &   93.10 &         7.47 &    0.00 &         0.00 &     1.72 &         0.00 \\
		&              & skilled &   29.31 &        22.99 &  100.00 &        21.84 &    0.57 &         0.00 &     2.87 &         0.57 \\
		&  RBF & random &    2.30 &         1.15 &   93.10 &        12.64 &    0.00 &         0.00 &     0.00 &         0.00 \\
		&              & skilled &   22.99 &        14.94 &  100.00 &        27.59 &    0.57 &         0.00 &     0.57 &         0.00 \\
		SigNet Madry &  Linear & random &    1.72 &         1.15 &   98.28 &        45.40 &    0.00 &         0.00 &     1.72 &         0.00 \\
		&              & skilled &   48.85 &        43.10 &  100.00 &        77.01 &    0.57 &         0.00 &     2.87 &         0.57 \\
		&  RBF & random &    2.30 &         1.72 &   97.70 &        62.64 &    0.00 &         0.00 &     0.00 &         0.00 \\
		&              & skilled &   45.98 &        40.23 &  100.00 &        84.48 &    0.57 &         0.00 &     0.57 &         0.00 \\
		\bottomrule
	\end{tabular}
\end{table*}

We now investigate the impact of simple noise-reduction techniques on the success of the attacks. Starting from the adversarial examples found in the experiments from the previous section, we applied the OTSU algorithm to remove noise with intensity lower than a threshold (as described in section \ref{sec:experimental}). We then evaluate if the resulting image remains adversarial. 

Tables \ref{tbl:otsu_evasion} and \ref{tbl:otsu_intrusion} evaluate the impact of processing the adversarial images with OTSU on the success rate of the attacks, for Type-I and Type-II attacks, respectively. We noticed that this pre-processing was effective against the gradient-free attacks, and provided some reduction in the success rate using gradient-based attacks. A possible explanation for this difference is that on the gradient-free methods used in these experiments, only small changes to a random set of pixels in done in each iteration, while the gradient-based methods can select larger changes to a smaller set of pixels (the regions where we have a large gradient of the loss w.r.t to the pixels).

\subsection{Limitations and practical considerations}
\label{sec:practical}

In this work we evaluated different attack scenarios (knowledge and capabilities for the attacker), but we would like to highlight some practical aspects to take into consideration for actual attacks:

\begin{itemize}
	\item \textbf{Discretization}: In this work, we use images in floating point representation, which is appropriate for the optimization methods. Images are commonly stored in 8-bits per channel (i.e. pixels intensities that are integer values $X_{ij} \in \{0,..., 255\}$). Simply rounding the pixel intensities to the nearest integer degrades the quality of adversarial examples. An alternative is to conduct a greedy search (changing each pixel at a time and checking if the image is still adversarial). This solution is computationally intensive, but can solve the problem (Carlini et al. \cite{carlini_towards_2017} reported  success with this search - i.e. by using this method, the discretized version of an adversarial image is still adversarial, for all images). For figures \ref{fig:cnn_attacks} and \ref{fig:lbp_attacks} we used the discretized images (and reported the score and RMSE using the discretized version of the images), so this step mainly adds more computational complexity for the attacker.

	\item \textbf{Physical Attacks}: We considered only attacks using digital images (i.e. after the sensor acquisition) which are limited for scenarios where digital images are used: services where the client provides a digital image (e.g. an app where the user scans a picture of a bank cheque). It has been shown that physical attacks are possible \cite{kurakin_adversarial_2016}, \cite{athalye_synthesizing_2017}, where adversarial images were printed, subsequently captured using a camera, and still fooled classifiers. However, this often requires more noise to be added, to account to transformations such as slight rotations or translations of the image. Also, it is worth noting that, if noise is printed on top of a handwritten signature, the noise $\delta$ needs to be constrained to be positive. In some early experiments in this scenario, we found it to also require more noise (50\% higher RMSE) than if $\delta$ does not have this constraint.

	\item \textbf{Knowledge of noise-removal}: In section \ref{sec:noise_removal}, we considered a pre-processing step to remove noise, that is effective (to some extent) in many scenarios. We note, however, that this cannot be considered a robust defense, and that if the adversary is aware of it, it can use this information as part of generating the adversarial images (e.g. knowing that a threshold $\tau$ is used, consider adding only pixels with intensity larger than $\tau$). This still increases the difficulty for gradient-based methods, since the problem becomes discontinuous (the pixel intensities can be 0 or greater than $\tau$).
\end{itemize}

\section{Conclusion}
\label{sec:conclusion}

In this paper we investigated the impact of adversarial examples on biometric systems, in particular by identifying threats to Offline Handwritten Signature Verification under the point of view of Adversarial Machine Learning. Our experiments indicate that the issue of adversarial examples present new threats to such systems in several scenarios, including both systems using handcrafted feature extractors and systems that learn directly from image pixels. 
In particular, we identify that Type-I attacks (changing a genuine signature so that it is rejected by the system) were successful is all systems investigated, even in a limited knowledge scenario, where the attacker does not have access to the signatures used for training the writer-dependent classifiers. The results in this scenario confirm previous findings that attacks transfer across different CNN classifiers \cite{szegedy_intriguing_2013}, and show that this transferability is also present on attacks on systems using a handcrafted feature extractor (CLBP). We found, however, that transferability is greatly reduced when the CNN is trained with a different set of \emph{users} (rather than a dijsoint set of \emph{samples} from the same classes, as investigated in \cite{szegedy_intriguing_2013}). 
We identified that Type-II attacks (changing a forgery to be accepted as genuine) are much harder to craft, obtaining lower success rates overall, and requiring larger amounts of noise for the strong gradient-based method. This contrasts with results in object recognition literature, where successful attacks (even in a targeted setting) are reported with much lower noise (less than 3 orders of magnitude), that are commonly visually imperceptible. \cite{carlini_towards_2017} 

Lastly, we investigated some countermeasures for this problem, and confirmed previous findings that the Madry defense \cite{madry_towards_2017} increase the amount of noise necessary to generate adversarial images. In this paper, we show that this defense is effective even when only applied on the feature learning phase, with no changes to the subsequent WD classifier training. We do note, however, that in spite of the increased amount of noise required, a strong attack (Carlini) is able to find adversarial examples most of the time. Our experiments with noise reduction show that this can reduce the success rate of attacks when the attacker is not aware of the defense, although we reiterate that this cannot be considered a robust defense (the adversary can incorporate this knowledge on the attack generation process). A definitive solution for this issue is yet an open research problem. Exploring the nature of the signal (a pen trajectory in 2D space) as part of the defense can be a promising direction for defenses. Another interesting area for future work is analyzing the impact of physical attacks (e.g. by printing adversarial noise on top of a signature).

\bibliographystyle{IEEEtran}
\bibliography{biblio}

\begin{IEEEbiography}[{\includegraphics[width=1in,height=1.25in,clip,keepaspectratio]{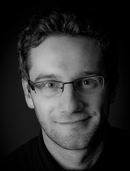}}]{Luiz G. Hafemann}
received his B.S. degree in Computer Science in 2008 and his M.Sc. degree in Informatics in 2014 from the Federal University of Paraná, Curitiba, PR, Brazil. He is currently pursuing a Ph.D. degree in the École de Technologie Supérieure, Université du Québec, in Montreal, QC, Canada. His research interests include Pattern Recognition, Machine Learning, Representation Learning and Handwritten Signature Verification.
\end{IEEEbiography}

\begin{IEEEbiography}[{\includegraphics[width=1in,height=1.25in,clip,keepaspectratio]{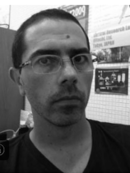}}]{Luiz S. Oliveira}
received his B.S. degree in Computer Science from Unicenp, Curitiba, PR, Brazil, the M.Sc. degree in electrical engineering and industrial informatics from the Centro Federal de Educação Tecnológica do Paraná (CEFET-PR), Curitiba, PR, Brazil, and Ph.D. degree in Computer Science from École de Technologie Supérieure, Université du Québec in 1995, 1998 and 2003, respectively. From 2004 to 2009 he was professor of the Computer Science Department at Pontifical Catholic University of Paraná, Curitiba, PR, Brazil. In 2009, he joined the Federal University of Paraná, Curitiba, PR, Brazil, where he is professor of the Department of Informatics and head of the Graduate Program in  Computer Science. His current interests include Pattern Recognition, Machine Learning, Image Analysis, and Evolutionary Computation.
\end{IEEEbiography}

\begin{IEEEbiography}[{\includegraphics[width=1in,height=1.25in,clip,keepaspectratio]{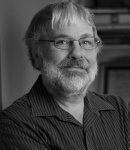}}]{Robert Sabourin}
joined the physics department of the Montreal University in 1977 where his main contribution was the design and the implementation of a microprocessor-based fine tracking system combined with a low-light level CCD detector. In 1983, he joined the staff of the École de Technologie Supérieure, Université du Québec, in Montreal, where he co-founded the Dept. of Automated Manufacturing Engineering where he is currently Full Professor and teaches Pattern Recognition, Evolutionary Algorithms, Neural Networks and Fuzzy Systems. In 1992, he joined also the Computer Science Department of the Pontificia Universidade Católica do Paraná (Curitiba, Brazil). Since 1996, he is a senior member of the Centre for Pattern Recognition and Machine Intelligence (CENPARMI, Concordia University). Since 2012, he is the Research Chair holder specializing in Adaptive Surveillance Systems in Dynamic Environments. Dr. Sabourin is the author (and co-author) of more than 400 scientific publications including journals and conference proceeding. His research interests are in the areas of adaptive biometric systems, adaptive classification systems in dynamic environments, dynamic classifier selection and evolutionary computation.

\end{IEEEbiography}

\clearpage
\renewcommand\appendixname{Supplementary material}

\appendix

\subsection{CNN architectures}

In this paper, we used the SigNet architecture \cite{hafemann_learning_2017} for the CNN-based experiments. This architecture is listed in Table \ref{table:signet}. Additionally, as base models for the Ensemble Adversarial Training \cite{tramer_ensemble_2017}, we trained two models based on similar architectures: SigNet-thin, that has a smaller amount of channels in the convolutional layers (described in Table \ref{table:signet_thin}) and SigNet-smaller that has less layers (described in Table \ref{table:signet_smaller}). In all cases, $M$ refer to the number of users (531 in the PK and LK1 experiments, and 264 in the LK2 experiments).

\begin{table}[h] \centering
	\caption{SigNet architecture}
	\resizebox{\columnwidth}{!}{%
		\begin{tabular}{@{}lcc@{}} \hline
			Layer & Size & Other Parameters \\ \hline
			
			Input & 1x150x220 \\
			Convolution (C1)& 96x11x11 & stride = 4, pad=0 \\
			Pooling & 96x3x3 & stride = 2\\
			
			Convolution (C2)& 256x5x5 & stride = 1, pad=2 \\
			Pooling & 256x3x3 & stride = 2\\
			
			Convolution (C3)& 384x3x3 & stride = 1, pad=1 \\
			Convolution (C4)& 384x3x3 & stride = 1, pad=1 \\
			Convolution (C5)& 256x3x3 & stride = 1, pad=1 \\
			Pooling & 256x3x3 & stride = 2 \\
			Fully Connected (FC6)& 2048 \\
			Fully Connected (FC7)& 2048 \\
			Fully Connected + Softmax & M & \\			
			\hline
		\end{tabular}
	}
	\label{table:signet}
\end{table}

\begin{table} \centering
	\caption{SigNet-thin architecture}
	\resizebox{\columnwidth}{!}{%
		\begin{tabular}{@{}lcc@{}} \hline
			Layer & Size & Other Parameters \\ \hline
			
			Input & 1x150x220 \\
			Convolution (C1)& 96x11x11 & stride = 4, pad=0 \\
			Pooling & 96x3x3 & stride = 2\\
			
			Convolution (C2)& 128x5x5 & stride = 1, pad=2 \\
			Pooling & 128x3x3 & stride = 2\\
			
			Convolution (C3)& 128x3x3 & stride = 1, pad=1 \\
			Convolution (C4)& 128x3x3 & stride = 1, pad=1 \\
			Convolution (C5)& 128x3x3 & stride = 1, pad=1 \\
			Pooling & 128x3x3 & stride = 2 \\
			Fully Connected (FC6)& 1024 \\
			Fully Connected (FC7)& 1024 \\
			Fully Connected + Softmax & M & \\
			\hline
		\end{tabular}
	}
	\label{table:signet_thin}
\end{table}

\begin{table} \centering
	\caption{SigNet-smaller architecture}
	\resizebox{\columnwidth}{!}{%
		\begin{tabular}{@{}lcc@{}} \hline
			Layer & Size & Other Parameters \\ \hline
			
			Input & 1x150x220 \\
			Convolution (C1)& 96x11x11 & stride = 4, pad=0 \\
			Pooling & 96x3x3 & stride = 2\\
			
			Convolution (C2)& 256x5x5 & stride = 1, pad=2 \\
			Pooling & 256x3x3 & stride = 2\\
			
			Convolution (C3)& 384x3x3 & stride = 1, pad=1 \\
			Convolution (C4)& 256x3x3 & stride = 1, pad=1 \\
			Pooling & 256x3x3 & stride = 2 \\
			Fully Connected (FC5)& 2048 \\
			Fully Connected + Softmax  & M & \\
			\hline
		\end{tabular}
	}
	\label{table:signet_smaller}
\end{table}

\subsection{Results on each dataset}
Results on the MCYT dataset for different knowledge scenarios are shown in Tables \ref{tbl:success_evasion_mcyt} to \ref{tbl:success_intrusion_lk2_mcyt}. Results with countermeasures are shown in tables \ref{tbl:success_defenses_evasion_mcyt} to \ref{tbl:mse_defenses_intrusion_mcyt}. Results with noise removal are shown in tables \ref{tbl:otsu_evasion_mcyt} and \ref{tbl:otsu_intrusion_mcyt}.

Results on the CEDAR dataset for different knowledge scenarios are shown in Tables \ref{tbl:success_evasion_cedar} to \ref{tbl:success_intrusion_lk2_cedar}. Results with countermeasures are shown in tables \ref{tbl:success_defenses_evasion_cedar} to \ref{tbl:mse_defenses_intrusion_cedar}. Results with noise removal are shown in tables \ref{tbl:otsu_evasion_cedar} and \ref{tbl:otsu_intrusion_cedar}.

Results on the Brazilian PUC-PR dataset for different knowledge scenarios are shown in Tables \ref{tbl:success_evasion_brazilian} to \ref{tbl:success_intrusion_lk2_brazilian}. Results with countermeasures are shown in tables \ref{tbl:success_defenses_evasion_brazilian} to \ref{tbl:mse_defenses_intrusion_brazilian}. Results with noise removal are shown in tables \ref{tbl:otsu_evasion_brazilian} and \ref{tbl:otsu_intrusion_brazilian}.

Results on the GPDS dataset for different knowledge scenarios are shown in Tables \ref{tbl:success_evasion_gpds} to \ref{tbl:success_intrusion_lk2_gpds}. Results with countermeasures are shown in tables \ref{tbl:success_defenses_evasion_gpds} to \ref{tbl:mse_defenses_intrusion_gpds}. Results with noise removal are shown in tables \ref{tbl:otsu_evasion_gpds} and \ref{tbl:otsu_intrusion_gpds}.

\begin{table}[h]
	\centering
	\caption{Success rate of Type-I attacks on the MCYT dataset(\% of attacks that transformed a genuine signature in a forgery)}
	\label{tbl:success_evasion_mcyt}
\begin{tabular}{ll|rrrr}
		\toprule
		& & \multicolumn{4}{c}{Attack Type} \\
		Feature &  Classifier &  FGM & Carlini & Anneal & Decision \\
		\midrule
		CLBP & Linear   &     - &        - &   55.56 &     75.00 \\
		CLBP & RBF      &     - &        - &  100.00 &    100.00 \\
		SigNet & Linear & 97.22 &   100.00 &   97.22 &    100.00 \\
		SigNet & RBF    & 91.67 &   100.00 &   97.22 &    100.00 \\
		
		\bottomrule
	\end{tabular}
\end{table}

\begin{table}
	\centering
	\caption{Distortion (RMSE of the adversarial noise) for successful Type-I attacks on the MCYT dataset}
	\label{tbl:mse_evasion_mcyt}
	\begin{tabular}{ll|rrrr}
		\toprule
		& & \multicolumn{4}{c}{Attack Type} \\
		Feature &  Classifier &  FGM & Carlini & Anneal & Decision \\
		\midrule
		CLBP & Linear   &    - &        - &    0.36 &      2.34 \\
		CLBP & RBF      &    - &        - &    0.36 &      $10^{-9}$ \\
		SigNet & Linear & 4.19 &     2.08 &    7.37 &      4.43 \\
		SigNet & RBF    & 4.22 &     2.17 &    6.80 &      4.16 \\
		
		\bottomrule
	\end{tabular}
\end{table}

\begin{table}
	\centering
	\caption{Success rate of Type-II attacks on the MCYT dataset (\% of attacks that transformed a forgery in a genuine signature)}
	\label{tbl:success_intrusion_mcyt}
	\resizebox{\columnwidth}{!}{%
		\begin{tabular}{lll|rrrr}
			\toprule
			&&& \multicolumn{4}{c}{Attack Type} \\
			&  &&   FGM &  Carlini &  Anneal &  Decision \\
			Features & Classifier & Forgery Type &       &          &         &           \\
			\midrule
			CLBP & Linear & random &     - &        - &   45.95 &     48.65 \\
			&& skilled &     - &        - &   45.95 &     48.65 \\
			CLBP & RBF & random &     - &        - &    0.00 &      0.00 \\
			&& skilled &     - &        - &    0.00 &      0.00 \\
			SigNet & Linear & random &  0.00 &   100.00 &    0.00 &      0.00 \\
			&& skilled & 40.54 &   100.00 &    2.70 &      2.70 \\
			SigNet & RBF & random &  0.00 &   100.00 &    0.00 &      0.00 \\
			&& skilled & 29.73 &   100.00 &    2.70 &      2.70 \\
			
			\bottomrule
		\end{tabular}
	}
	
\end{table}

\begin{table}
	\centering
	\caption{Distortion (RMSE of the adversarial noise) for successful Type-II attacks on the MCYT dataset}
	\label{tbl:mse_intrusion_mcyt}
	\resizebox{\columnwidth}{!}{%
		\begin{tabular}{lll|rrrr}
			\toprule
			&&& \multicolumn{4}{c}{Attack Type} \\
			&  &&   FGM &  Carlini &  Anneal &  Decision \\
			Features & Classifier & Forgery Type &       &          &         &           \\
			\midrule
			CLBP & Linear & random &    - &        - &    0.45 &      0.47 \\
			&& skilled &    - &        - &    0.53 &      0.67 \\
			SigNet & Linear & random &    - &     7.78 &       - &         - \\
			&& skilled & 4.29 &     4.13 &    1.72 &      0.52 \\
			SigNet & RBF & random &    - &     8.53 &       - &         - \\
			&& skilled & 4.24 &     4.54 &    0.94 &      0.27 \\
			
			\bottomrule
		\end{tabular}
	}
\end{table}

\begin{table}
	\centering
	\caption{Success rate of Type-I attacks on the MCYT dataset (\% of attacks that transformed a genuine signature in a forgery)  (Limited Knowledge)}
	\label{tbl:success_evasion_lk1_mcyt}
	\begin{tabular}{ll|rrrr}
		\toprule
		& & \multicolumn{4}{c}{Attack Type} \\
		Feature &  Classifier &  FGM & Carlini & Anneal & Decision \\
		\midrule
		CLBP & Linear   &     - &        - &   38.89 &     41.67 \\
		CLBP & RBF      &     - &        - &   77.78 &     77.78 \\
		SigNet & Linear & 88.89 &    80.56 &   52.78 &     44.44 \\
		SigNet & RBF    & 88.89 &    86.11 &   55.56 &     50.00 \\
		
		\bottomrule
	\end{tabular}
\end{table}

\begin{table}
	\centering
	\caption{Success rate of Type-II attacks on the MCYT dataset (\% of attacks that transformed a forgery in a genuine signature)  (Limited Knowledge)}
	\label{tbl:success_intrusion_lk1_mcyt}
	\resizebox{\columnwidth}{!}{%
		\begin{tabular}{lll|rrrr}
			\toprule
			&&& \multicolumn{4}{c}{Attack Type} \\
			&  &&   FGM &  Carlini &  Anneal &  Decision \\
			Features & Classifier & Forgery Type &       &          &         &           \\
			\midrule
			
			CLBP & Linear & random &     - &        - &   29.73 &     35.14 \\
			&& skilled &     - &        - &   27.03 &     29.73 \\
			CLBP & RBF & random &     - &        - &    0.00 &      0.00 \\
			&& skilled &     - &        - &    0.00 &      0.00 \\
			SigNet & Linear & random &  0.00 &    54.05 &    0.00 &      0.00 \\
			&& skilled & 21.62 &    78.38 &    2.70 &      0.00 \\
			SigNet & RBF & random &  0.00 &    78.38 &    0.00 &      0.00 \\
			&& skilled & 24.32 &    86.49 &    2.70 &      0.00 \\

			\bottomrule
		\end{tabular}
	}	
\end{table}

\begin{table}
	\centering
	\caption{Success rate of Type-I attacks on the MCYT dataset (\% of attacks that transformed a genuine signature in a forgery)  (Limited Knowledge \#2)}
	\label{tbl:success_evasion_lk2_mcyt}
	\begin{tabular}{ll|rrrr}
		\toprule
		& & \multicolumn{4}{c}{Attack Type} \\
		Feature &  Classifier &  FGM & Carlini & Anneal & Decision \\
		\midrule
		SigNet & Linear & 37.84 &     8.11 &   40.54 &     18.92 \\
		SigNet & RBF    & 43.24 &    10.81 &   40.54 &     18.92 \\
		
		\bottomrule
	\end{tabular}
\end{table}

\begin{table}
	\centering
	\caption{Success rate of Type-II attacks on the MCYT dataset (\% of attacks that transformed a forgery in a genuine signature)  (Limited Knowledge \#2)}
	\label{tbl:success_intrusion_lk2_mcyt}
	\resizebox{\columnwidth}{!}{%
		\begin{tabular}{lll|rrrr}
			\toprule
			&&& \multicolumn{4}{c}{Attack Type} \\
			&  &&   FGM &  Carlini &  Anneal &  Decision \\
			Features & Classifier & Forgery Type &       &          &         &           \\
			\midrule
			SigNet & Linear & random & 0.00 &     0.00 &    0.00 &      0.00 \\
			&& skilled & 0.00 &     2.70 &    0.00 &      0.00 \\
			SigNet & RBF & random & 0.00 &     0.00 &    0.00 &      0.00 \\
			&& skilled & 5.41 &     2.70 &    0.00 &      0.00 \\		
			\bottomrule
		\end{tabular}
	}	
\end{table}

\begin{table}
	\caption{Success rate of Type-I attacks on the MCYT dataset considering different defenses and attacker knowledge scenarios}
	\label{tbl:success_defenses_evasion_mcyt}
	\resizebox{\columnwidth}{!}{%
		\begin{tabular}{ll|rrr|rrr}
			\toprule
			&& \multicolumn{6}{c}{Attack Type and Knowledge scenario} \\
			&  &   \multicolumn{3}{c}{FGM} & \multicolumn{3}{c}{Carlini}\\
			Defense & Classifier & PK &   LK1 &   LK2 &     PK &   LK1 &  LK2 \\
			\midrule
			Baseline & Linear & 100.00 & 86.49 & 37.84 &  100.00 & 75.68 &  8.11 \\
			&  RBF & 100.00 & 89.19 & 43.24 &  100.00 & 81.08 & 10.81 \\
			Ens. Adv. & Linear &  70.27 & 64.86 & 16.22 &  100.00 & 89.19 &  8.11 \\
			& RBF &  72.97 & 59.46 & 16.22 &  100.00 & 86.49 &  5.41 \\
			Madry & Linear &  67.57 & 45.95 &  8.11 &  100.00 & 78.38 &  2.70 \\
			& RBF &  56.76 & 48.65 &  8.11 &  100.00 & 86.49 &  0.00 \\
			
			\bottomrule
		\end{tabular}
	}
\end{table}

\begin{table}
	\caption{Distortion (RMSE of the adversarial noise) for Type-I attacks on the MCYT dataset, considering different defenses and attacker knowledge scenarios}
	\label{tbl:mse_defenses_evasion_mcyt}
	\resizebox{\columnwidth}{!}{%
		\begin{tabular}{ll|rrr|rrr}
			\toprule
			&& \multicolumn{6}{c}{Attack Type and Knowledge scenario} \\
			&  &   \multicolumn{3}{c}{FGM} & \multicolumn{3}{c}{Carlini}\\
			Defense & Classifier & PK &   LK1 &   LK2 &     PK &   LK1 &  LK2 \\
			\midrule
			baseline & Linear & 4.25 & 4.29 & 4.46 &    2.03 & 2.05 & 2.22 \\
			& RBF & 4.29 & 4.32 & 4.49 &    2.16 & 2.15 & 2.52 \\
			ensadv & Linear & 4.58 & 4.52 & 4.38 &    2.19 & 2.33 & 2.05 \\
			& RBF & 4.55 & 4.50 & 4.34 &    2.35 & 2.48 & 2.23 \\
			madry & Linear & 4.88 & 4.90 & 4.05 &    4.94 & 4.96 & 1.95 \\
			& RBF & 4.87 & 4.90 & 4.07 &    5.26 & 5.54 &    - \\
			\bottomrule
		\end{tabular}
	}
\end{table}

\begin{table*}
	\centering
	\caption{Success rate of Type-II attacks on the MCYT dataset considering different defenses and attacker knowledge scenarios}
	\label{tbl:success_defenses_intrusion_mcyt}
	
	\begin{tabular}{lll|rrr|rrr}
		\toprule
		&&& \multicolumn{6}{c}{Attack Type and Knowledge scenario} \\
		&&  &   \multicolumn{3}{c}{FGM} & \multicolumn{3}{c}{Carlini}\\
		Defense & Classifier & Forgery Type & PK &   LK1 &   LK2 &     PK &   LK1 &  LK2 \\
		\midrule
		baseline & Linear & random &  0.00 &  0.00 & 0.00 &  100.00 & 51.35 & 0.00 \\
		&              & skilled & 35.14 & 32.43 & 0.00 &  100.00 & 67.57 & 2.70 \\
		& RBF & random &  0.00 &  0.00 & 0.00 &  100.00 & 70.27 & 0.00 \\
		&              & skilled & 35.14 & 29.73 & 5.41 &  100.00 & 83.78 & 2.70 \\
		ensadv & Linear & random &  0.00 &  0.00 & 0.00 &   97.30 & 45.95 & 0.00 \\
		&              & skilled & 29.73 & 13.51 & 0.00 &  100.00 & 78.38 & 0.00 \\
		& RBF & random &  0.00 &  0.00 & 0.00 &   97.30 & 70.27 & 0.00 \\
		&              & skilled & 24.32 & 21.62 & 0.00 &  100.00 & 83.78 & 0.00 \\
		madry & Linear & random &  0.00 &  0.00 & 0.00 &  100.00 & 54.05 & 0.00 \\
		&              & skilled & 35.14 & 27.03 & 2.70 &  100.00 & 78.38 & 2.70 \\
		& RBF & random &  0.00 &  0.00 & 0.00 &  100.00 & 78.38 & 0.00 \\
		&              & skilled & 35.14 & 27.03 & 5.41 &  100.00 & 94.59 & 5.41 \\
		
		\bottomrule
	\end{tabular}
\end{table*}

\begin{table*}
	\centering
	\caption{Distortion (RMSE of the adversarial noise) for Type-II attacks on the MCYT dataset, considering different defenses and attacker knowledge scenarios}
	\label{tbl:mse_defenses_intrusion_mcyt}
	\begin{tabular}{lll|rrr|rrr}
		\toprule
		&&& \multicolumn{6}{c}{Attack Type and Knowledge scenario} \\
		&&  &   \multicolumn{3}{c}{FGM} & \multicolumn{3}{c}{Carlini}\\
		Defense & Classifier & Forgery Type & PK &   LK1 &   LK2 &     PK &   LK1 &  LK2 \\
		\midrule
		baseline & Linear & random &    - &    - &    - &    7.02 &  7.00 &    - \\
		&              & skilled & 4.41 & 4.42 &    - &    3.69 &  3.97 & 3.56 \\
		& RBF & random &    - &    - &    - &    7.89 &  8.28 &    - \\
		&              & skilled & 4.39 & 4.45 & 4.75 &    4.22 &  4.39 & 2.26 \\
		ensadv & Linear & random &    - &    - &    - &   12.41 & 11.93 &    - \\
		&              & skilled & 4.49 & 4.41 &    - &    6.77 &  7.11 &    - \\
		& RBF & random &    - &    - &    - &   13.05 & 13.66 &    - \\
		&              & skilled & 4.51 & 4.49 &    - &    7.68 &  7.23 &    - \\
		madry & Linear & random &    - &    - &    - &   15.01 & 15.83 &    - \\
		&              & skilled & 4.91 & 4.99 & 4.31 &    8.91 &  9.16 & 3.73 \\
		& RBF & random &    - &    - &    - &   15.79 & 16.14 &    - \\
		&              & skilled & 4.94 & 5.00 & 4.10 &    9.85 &  9.90 & 3.68 \\
		\bottomrule
	\end{tabular}
\end{table*}

\begin{table*}
	\centering
	\caption{Success of Type-I attacks on the MCYT dataset in a PK scenario, with no pre-processing and with OTSU pre-processing}
	\label{tbl:otsu_evasion_mcyt}
	
	\begin{tabular}{ll|rrrrrrrr}
		\toprule
		&& \multicolumn{8}{c}{Attack Type and Preprocessing} \\
		& & \multicolumn{2}{c}{FGM} & \multicolumn{2}{c}{Carlini} & \multicolumn{2}{l}{Anneal} & \multicolumn{2}{c}{Decision} \\
		&  & None & OTSU & None & OTSU & None & OTSU & None & OTSU\\
		Feature & Classifier  &          &              &         &              &         &              &          &              \\
		\midrule
		CLBP & Linear &       - &            - &       - &            - &   55.56 &         8.33 &    75.00 &         2.78 \\
		& RBF &       - &            - &       - &            - &  100.00 &         0.00 &   100.00 &         0.00 \\
		SigNet Baseline & Linear &  100.00 &        67.57 &  100.00 &        21.62 &   94.59 &         0.00 &   100.00 &         0.00 \\
		& RBF &  100.00 &        62.16 &  100.00 &        24.32 &   97.30 &         0.00 &   100.00 &         0.00 \\
		SigNet Ens. Adv. & Linear &   70.27 &        43.24 &  100.00 &         0.00 &   91.89 &         0.00 &   100.00 &         0.00 \\
		& RBF &   72.97 &        35.14 &  100.00 &         0.00 &   94.59 &         0.00 &   100.00 &         0.00 \\
		SigNet Madry & Linear &   67.57 &        59.46 &  100.00 &        56.76 &   64.86 &         0.00 &    97.30 &         2.70 \\
		& RBF &   56.76 &        51.35 &  100.00 &        59.46 &   67.57 &         0.00 &   100.00 &         2.70 \\
		
		\bottomrule
	\end{tabular}
\end{table*}

\begin{table*}
	\centering
	\caption{Success of Type-II attacks on the MCYT dataset in a PK scenario, with no pre-processing and with OTSU pre-processing}
	\label{tbl:otsu_intrusion_mcyt}
	\begin{tabular}{lll|rrrrrrrr}
		\toprule
		&&& \multicolumn{8}{c}{Attack Type and Preprocessing} \\
		&& & \multicolumn{2}{c}{FGM} & \multicolumn{2}{c}{Carlini} & \multicolumn{2}{l}{Anneal} & \multicolumn{2}{c}{Decision} \\
		&&  & None & OTSU & None & OTSU & None & OTSU & None & OTSU\\
		Feature & Classifier  & Forgery Type &         &              &         &              &         &              &          &              \\
		\midrule
		CLBP & Linear & random &       - &            - &       - &            - &   45.95 &         0.00 &    48.65 &         0.00 \\
		&              & skilled &       - &            - &       - &            - &   45.95 &         0.00 &    48.65 &         2.70 \\
		& RBF & random &       - &            - &       - &            - &    0.00 &         0.00 &     0.00 &         0.00 \\
		&              & skilled &       - &            - &       - &            - &    0.00 &         0.00 &     0.00 &         0.00 \\
\bottomrule
	\end{tabular}
\end{table*}

%
%
%

\begin{table}
	\centering
	\caption{Success rate of Type-I attacks on the CEDAR dataset(\% of attacks that transformed a genuine signature in a forgery)}
	\label{tbl:success_evasion_cedar}
	\begin{tabular}{ll|rrrr}
		\toprule
		& & \multicolumn{4}{c}{Attack Type} \\
		Feature &  Classifier &  FGM & Carlini & Anneal & Decision \\
		\midrule
		CLBP & Linear   &      - &        - &   70.37 &     85.19 \\
		CLBP & RBF      &      - &        - &  100.00 &    100.00 \\
		SigNet & Linear & 100.00 &   100.00 &  100.00 &    100.00 \\
		SigNet & RBF    & 100.00 &   100.00 &  100.00 &    100.00 \\

		\bottomrule
	\end{tabular}
\end{table}

\begin{table}
	\centering
	\caption{Distortion (RMSE of the adversarial noise) for successful Type-I attacks on the CEDAR dataset}
	\label{tbl:mse_evasion_cedar}
	\begin{tabular}{ll|rrrr}
		\toprule
		& & \multicolumn{4}{c}{Attack Type} \\
		Feature &  Classifier &  FGM & Carlini & Anneal & Decision \\
		\midrule
		CLBP & Linear   &    - &        - &    0.36 &      0.69 \\
		CLBP & RBF      &    - &        - &    0.36 &      $10^{-9}$ \\
		SigNet & Linear & 3.83 &     0.85 &    4.69 &      2.43 \\
		SigNet & RBF    & 3.81 &     0.88 &    3.99 &      2.09 \\

		\bottomrule
	\end{tabular}
\end{table}

\begin{table}
	\centering
	\caption{Success rate of Type-II attacks on the CEDAR dataset (\% of attacks that transformed a forgery in a genuine signature)}
	\label{tbl:success_intrusion_cedar}
	\resizebox{\columnwidth}{!}{%
		\begin{tabular}{lll|rrrr}
			\toprule
			&&& \multicolumn{4}{c}{Attack Type} \\
			&  &&   FGM &  Carlini &  Anneal &  Decision \\
			Features & Classifier & Forgery Type &       &          &         &           \\
			\midrule
			CLBP & Linear & random &     - &        - &   25.93 &     44.44 \\
			&& skilled &     - &        - &   34.62 &     46.15 \\
			CLBP & RBF & random &     - &        - &    0.00 &      0.00 \\
			&& skilled &     - &        - &    0.00 &      0.00 \\
			SigNet & Linear & random &  0.00 &   100.00 &    0.00 &      0.00 \\
			&& skilled & 30.77 &   100.00 &    3.85 &      7.69 \\
			SigNet & RBF & random &  0.00 &   100.00 &    0.00 &      0.00 \\
			&& skilled & 26.92 &   100.00 &    0.00 &      0.00 \\

			\bottomrule
		\end{tabular}
	}
	
\end{table}

\begin{table}
	\centering
	\caption{Distortion (RMSE of the adversarial noise) for successful Type-II attacks on the CEDAR dataset}
	\label{tbl:mse_intrusion_cedar}
	\resizebox{\columnwidth}{!}{%
		\begin{tabular}{lll|rrrr}
			\toprule
			&&& \multicolumn{4}{c}{Attack Type} \\
			&  &&   FGM &  Carlini &  Anneal &  Decision \\
			Features & Classifier & Forgery Type &       &          &         &           \\
			\midrule
			CLBP & Linear & random &    - &        - &    0.35 &      0.72 \\
			&& skilled &    - &        - &    0.38 &      1.12 \\
			SigNet & Linear & random &    - &     3.38 &       - &         - \\
			&& skilled & 3.96 &     2.16 &    1.30 &      0.88 \\
			SigNet & RBF & random &    - &     4.05 &       - &         - \\
			&& skilled & 3.73 &     2.58 &       - &         - \\
			\bottomrule
		\end{tabular}
	}
\end{table}

\begin{table}
	\centering
	\caption{Success rate of Type-I attacks on the CEDAR dataset (\% of attacks that transformed a genuine signature in a forgery)  (Limited Knowledge)}
	\label{tbl:success_evasion_lk1_cedar}
	\begin{tabular}{ll|rrrr}
		\toprule
		& & \multicolumn{4}{c}{Attack Type} \\
		Feature &  Classifier &  FGM & Carlini & Anneal & Decision \\
		\midrule		
		CLBP & Linear   &      - &        - &   33.33 &     33.33 \\
		CLBP & RBF      &      - &        - &   85.19 &     85.19 \\
		SigNet & Linear & 100.00 &    77.78 &   62.96 &     40.74 \\
		SigNet & RBF    & 100.00 &    96.30 &   62.96 &     48.15 \\
		
		\bottomrule
	\end{tabular}
\end{table}

\begin{table}
	\centering
	\caption{Success rate of Type-II attacks on the CEDAR dataset (\% of attacks that transformed a forgery in a genuine signature)  (Limited Knowledge)}
	\label{tbl:success_intrusion_lk1_cedar}
	\resizebox{\columnwidth}{!}{%
		\begin{tabular}{lll|rrrr}
			\toprule
			&&& \multicolumn{4}{c}{Attack Type} \\
			&  &&   FGM &  Carlini &  Anneal &  Decision \\
			Features & Classifier & Forgery Type &       &          &         &           \\
			\midrule
			
			CLBP & Linear & random &     - &        - &   18.52 &     14.81 \\
			&& skilled &     - &        - &   15.38 &     15.38 \\
			CLBP & RBF & random &     - &        - &    0.00 &      0.00 \\
			&& skilled &     - &        - &    0.00 &      0.00 \\
			SigNet & Linear & random &  0.00 &    37.04 &    0.00 &      0.00 \\
			&& skilled & 30.77 &    65.38 &    3.85 &      0.00 \\
			SigNet & RBF & random &  0.00 &    70.37 &    0.00 &      0.00 \\
			&& skilled & 23.08 &    76.92 &    0.00 &      0.00 \\

			\bottomrule
		\end{tabular}
	}	
\end{table}

\begin{table}
	\centering
	\caption{Success rate of Type-I attacks on the CEDAR dataset (\% of attacks that transformed a genuine signature in a forgery)  (Limited Knowledge \#2)}
	\label{tbl:success_evasion_lk2_cedar}
	\begin{tabular}{ll|rrrr}
		\toprule
		& & \multicolumn{4}{c}{Attack Type} \\
		Feature &  Classifier &  FGM & Carlini & Anneal & Decision \\
		\midrule
		SigNet & Linear & 70.37 &     3.70 &   55.56 &     25.93 \\
		SigNet & RBF    & 74.07 &    11.11 &   59.26 &     22.22 \\

		\bottomrule
	\end{tabular}
\end{table}

\begin{table}
	\centering
	\caption{Success rate of Type-II attacks on the CEDAR dataset (\% of attacks that transformed a forgery in a genuine signature)  (Limited Knowledge \#2)}
	\label{tbl:success_intrusion_lk2_cedar}
	\resizebox{\columnwidth}{!}{%
		\begin{tabular}{lll|rrrr}
			\toprule
			&&& \multicolumn{4}{c}{Attack Type} \\
			&  &&   FGM &  Carlini &  Anneal &  Decision \\
			Features & Classifier & Forgery Type &       &          &         &           \\
			\midrule
			SigNet & Linear & random & 0.00 &     0.00 &    0.00 &      0.00 \\
			&& skilled & 0.00 &     0.00 &    3.70 &      3.70 \\
			SigNet & RBF & random & 0.00 &     0.00 &    0.00 &      0.00 \\
			&& skilled & 0.00 &     3.70 &    7.41 &      0.00 \\
			
			\bottomrule
		\end{tabular}
	}	
\end{table}

\begin{table}
	\caption{Success rate of Type-I attacks on the CEDAR dataset considering different defenses and attacker knowledge scenarios}
	\label{tbl:success_defenses_evasion_cedar}
	\resizebox{\columnwidth}{!}{%
		\begin{tabular}{ll|rrr|rrr}
			\toprule
			&& \multicolumn{6}{c}{Attack Type and Knowledge scenario} \\
			&  &   \multicolumn{3}{c}{FGM} & \multicolumn{3}{c}{Carlini}\\
			Defense & Classifier & PK &   LK1 &   LK2 &     PK &   LK1 &  LK2 \\
			\midrule
			
			baseline & Linear & 100.00 &  96.30 & 70.37 &  100.00 & 74.07 &  3.70 \\
			& RBF & 100.00 & 100.00 & 74.07 &  100.00 & 81.48 & 11.11 \\
			ensadv & Linear & 100.00 & 100.00 & 70.37 &  100.00 & 77.78 &  3.70 \\
			& RBF & 100.00 & 100.00 & 70.37 &  100.00 & 96.30 &  3.70 \\
			madry & Linear & 100.00 &  96.30 & 33.33 &  100.00 & 77.78 &  0.00 \\
			& RBF & 100.00 & 100.00 & 25.93 &  100.00 & 92.59 &  0.00 \\

			\bottomrule
		\end{tabular}
	}
\end{table}

\begin{table}
	\caption{Distortion (RMSE of the adversarial noise) for Type-I attacks on the CEDAR dataset, considering different defenses and attacker knowledge scenarios}
	\label{tbl:mse_defenses_evasion_cedar}
	\resizebox{\columnwidth}{!}{%
		\begin{tabular}{ll|rrr|rrr}
			\toprule
			&& \multicolumn{6}{c}{Attack Type and Knowledge scenario} \\
			&  &   \multicolumn{3}{c}{FGM} & \multicolumn{3}{c}{Carlini}\\
			Defense & Classifier & PK &   LK1 &   LK2 &     PK &   LK1 &  LK2 \\
			\midrule
			baseline & Linear & 3.93 & 3.99 & 4.07 &    0.86 & 0.91 & 0.85 \\
			& RBF & 3.96 & 3.97 & 4.07 &    0.94 & 0.92 & 0.97 \\
			ensadv & Linear & 4.22 & 4.09 & 4.03 &    0.82 & 0.86 & 1.10 \\
			& RBF & 4.20 & 4.15 & 4.00 &    0.90 & 0.84 & 1.02 \\
			madry & Linear & 4.49 & 4.43 & 4.14 &    2.01 & 2.10 &    - \\
			& RBF & 4.54 & 4.45 & 4.31 &    2.34 & 2.30 &    - \\
			\bottomrule
		\end{tabular}
	}
\end{table}

\begin{table*}
	\centering
	\caption{Success rate of Type-II attacks on the CEDAR dataset considering different defenses and attacker knowledge scenarios}
	\label{tbl:success_defenses_intrusion_cedar}
	
	\begin{tabular}{lll|rrr|rrr}
		\toprule
		&&& \multicolumn{6}{c}{Attack Type and Knowledge scenario} \\
		&&  &   \multicolumn{3}{c}{FGM} & \multicolumn{3}{c}{Carlini}\\
		Defense & Classifier & Forgery Type & PK &   LK1 &   LK2 &     PK &   LK1 &  LK2 \\
		\midrule
		baseline & Linear & random & 11.11 &  7.41 &  0.00 &  100.00 & 37.04 & 0.00 \\
		&              & skilled & 48.15 & 25.93 &  0.00 &  100.00 & 48.15 & 0.00 \\
		& RBF & random & 11.11 &  7.41 &  0.00 &  100.00 & 70.37 & 0.00 \\
		&              & skilled & 40.74 & 37.04 &  0.00 &  100.00 & 74.07 & 3.70 \\
		ensadv & Linear & random &  7.41 &  0.00 &  0.00 &  100.00 & 40.74 & 0.00 \\
		&              & skilled & 29.63 & 11.11 &  0.00 &  100.00 & 62.96 & 7.41 \\
		& RBF & random & 11.11 &  0.00 &  0.00 &  100.00 & 66.67 & 0.00 \\
		&              & skilled & 22.22 & 14.81 &  0.00 &  100.00 & 77.78 & 3.70 \\
		madry & Linear & random &  7.41 &  0.00 &  0.00 &  100.00 & 40.74 & 0.00 \\
		&              & skilled & 59.26 & 44.44 & 22.22 &  100.00 & 62.96 & 7.41 \\
		& RBF & random &  3.70 &  0.00 &  0.00 &  100.00 & 81.48 & 0.00 \\
		&              & skilled & 55.56 & 37.04 & 11.11 &  100.00 & 81.48 & 7.41 \\
		
		\bottomrule
	\end{tabular}
\end{table*}

\begin{table*}
	\centering
	\caption{Distortion (RMSE of the adversarial noise) for Type-II attacks on the CEDAR dataset, considering different defenses and attacker knowledge scenarios}
	\label{tbl:mse_defenses_intrusion_cedar}
	\begin{tabular}{lll|rrr|rrr}
		\toprule
		&&& \multicolumn{6}{c}{Attack Type and Knowledge scenario} \\
		&&  &   \multicolumn{3}{c}{FGM} & \multicolumn{3}{c}{Carlini}\\
		Defense & Classifier & Forgery Type & PK &   LK1 &   LK2 &     PK &   LK1 &  LK2 \\
		\midrule
		baseline & Linear & random & 3.93 & 3.75 &    - &    3.50 & 4.49 &    - \\
		&              & skilled & 4.03 & 3.85 &    - &    2.04 & 1.60 &    - \\
		& RBF & random & 3.84 & 3.83 &    - &    4.11 & 4.24 &    - \\
		&              & skilled & 3.97 & 3.93 &    - &    2.40 & 2.33 & 0.77 \\
		ensadv & Linear & random & 4.43 &    - &    - &    4.82 & 5.00 &    - \\
		&              & skilled & 4.40 & 4.65 &    - &    3.13 & 2.47 & 1.11 \\
		& RBF & random & 4.28 &    - &    - &    5.87 & 4.83 &    - \\
		&              & skilled & 4.46 & 4.46 &    - &    3.76 & 3.13 & 1.82 \\
		madry & Linear & random & 4.55 &    - &    - &    5.66 & 5.73 &    - \\
		&              & skilled & 4.88 & 5.00 & 4.26 &    3.71 & 3.20 & 1.32 \\
		& RBF & random & 4.09 &    - &    - &    6.45 & 6.34 &    - \\
		&              & skilled & 4.87 & 4.94 & 4.26 &    4.71 & 4.42 & 1.47 \\
		\bottomrule
	\end{tabular}
\end{table*}

\begin{table*}
	\centering
	\caption{Success of Type-I attacks on the CEDAR dataset in a PK scenario, with no pre-processing and with OTSU pre-processing}
	\label{tbl:otsu_evasion_cedar}
	
	\begin{tabular}{ll|rrrrrrrr}
		\toprule
		&& \multicolumn{8}{c}{Attack Type and Preprocessing} \\
		& & \multicolumn{2}{c}{FGM} & \multicolumn{2}{c}{Carlini} & \multicolumn{2}{l}{Anneal} & \multicolumn{2}{c}{Decision} \\
		&  & None & OTSU & None & OTSU & None & OTSU & None & OTSU\\
		Feature & Classifier  &          &              &         &              &         &              &          &              \\
		\midrule
		CLBP & Linear &       - &            - &       - &            - &   70.37 &         7.41 &    85.19 &         0.00 \\
		& RBF &       - &            - &       - &            - &  100.00 &         0.00 &   100.00 &         0.00 \\
		SigNet Baseline & Linear &  100.00 &       100.00 &  100.00 &        11.11 &  100.00 &         0.00 &   100.00 &         0.00 \\
		& RBF &  100.00 &       100.00 &  100.00 &        14.81 &   96.30 &         0.00 &   100.00 &         0.00 \\
		SigNet Ens. Adv. & Linear &  100.00 &        81.48 &  100.00 &         0.00 &  100.00 &         0.00 &   100.00 &         0.00 \\
		& RBF &  100.00 &        81.48 &  100.00 &         0.00 &  100.00 &         0.00 &   100.00 &         0.00 \\
		SigNet Madry & Linear &  100.00 &       100.00 &  100.00 &        85.19 &  100.00 &         0.00 &   100.00 &         3.70 \\
		& RBF &  100.00 &       100.00 &  100.00 &        81.48 &  100.00 &         0.00 &   100.00 &         0.00 \\

		\bottomrule
	\end{tabular}
\end{table*}

\begin{table*}
	\centering
	\caption{Success of Type-II attacks on the CEDAR dataset in a PK scenario, with no pre-processing and with OTSU pre-processing}
	\label{tbl:otsu_intrusion_cedar}
	\begin{tabular}{lll|rrrrrrrr}
		\toprule
		&&& \multicolumn{8}{c}{Attack Type and Preprocessing} \\
		&& & \multicolumn{2}{c}{FGM} & \multicolumn{2}{c}{Carlini} & \multicolumn{2}{l}{Anneal} & \multicolumn{2}{c}{Decision} \\
		&&  & None & OTSU & None & OTSU & None & OTSU & None & OTSU\\
		Feature & Classifier  & Forgery Type &         &              &         &              &         &              &          &              \\
		\midrule
		CLBP & Linear & random &       - &            - &       - &            - &   25.93 &         0.00 &    44.44 &         0.00 \\
		&              & skilled &       - &            - &       - &            - &   34.62 &         0.00 &    46.15 &         3.85 \\
		& RBF & random &       - &            - &       - &            - &    0.00 &         0.00 &     0.00 &         0.00 \\
		&              & skilled &       - &            - &       - &            - &    0.00 &         0.00 &     0.00 &         0.00 \\
\bottomrule
	\end{tabular}
\end{table*}

%
%

\begin{table}
	\centering
	\caption{Success rate of Type-I attacks on the Brazilian PUC-PR dataset(\% of attacks that transformed a genuine signature in a forgery)}
	\label{tbl:success_evasion_brazilian}
\begin{tabular}{ll|rrrr}
		\toprule
		& & \multicolumn{4}{c}{Attack Type} \\
		Feature &  Classifier &  FGM & Carlini & Anneal & Decision \\
		\midrule
		CLBP & Linear   &      - &        - &   66.67 &     76.67 \\
		CLBP & RBF      &      - &        - &  100.00 &    100.00 \\
		SigNet & Linear & 100.00 &   100.00 &  100.00 &    100.00 \\
		SigNet & RBF    & 100.00 &   100.00 &  100.00 &    100.00 \\
		
		\bottomrule
	\end{tabular}
\end{table}

\begin{table}
	\centering
	\caption{Distortion (RMSE of the adversarial noise) for successful Type-I attacks on the Brazilian PUC-PR dataset}
	\label{tbl:mse_evasion_brazilian}
	\begin{tabular}{ll|rrrr}
		\toprule
		& & \multicolumn{4}{c}{Attack Type} \\
		Feature &  Classifier &  FGM & Carlini & Anneal & Decision \\
		\midrule
		
		CLBP & Linear   &    - &        - &    0.36 &      0.37 \\
		CLBP & RBF      &    - &        - &    0.36 &      $10^{-9}$ \\
		SigNet & Linear & 3.92 &     1.03 &    5.10 &      2.75 \\
		SigNet & RBF    & 3.89 &     1.05 &    4.46 &      2.51 \\

		\bottomrule
	\end{tabular}
\end{table}

\begin{table}
	\centering
	\caption{Success rate of Type-II attacks on the Brazilian PUC-PR dataset (\% of attacks that transformed a forgery in a genuine signature)}
	\label{tbl:success_intrusion_brazilian}
	\resizebox{\columnwidth}{!}{%
		\begin{tabular}{lll|rrrr}
			\toprule
			&&& \multicolumn{4}{c}{Attack Type} \\
			&  &&   FGM &  Carlini &  Anneal &  Decision \\
			Features & Classifier & Forgery Type &       &          &         &           \\
			\midrule
			CLBP & Linear & random &     - &        - &   33.33 &     36.67 \\
			&& skilled &     - &        - &   33.33 &     36.67 \\
			CLBP & RBF & random &     - &        - &    0.00 &      0.00 \\
			&& skilled &     - &        - &    0.00 &      0.00 \\
			SigNet & Linear & random &  6.67 &   100.00 &    0.00 &      0.00 \\
			&& skilled & 33.33 &   100.00 &    0.00 &      3.33 \\
			SigNet & RBF & random &  3.33 &   100.00 &    0.00 &      0.00 \\
			&& skilled & 10.00 &   100.00 &    3.33 &      3.33 \\

			\bottomrule
		\end{tabular}
	}
	
\end{table}

\begin{table}
	\centering
	\caption{Distortion (RMSE of the adversarial noise) for successful Type-II attacks on the Brazilian PUC-PR dataset}
	\label{tbl:mse_intrusion_brazilian}
	\resizebox{\columnwidth}{!}{%
		\begin{tabular}{lll|rrrr}
			\toprule
			&&& \multicolumn{4}{c}{Attack Type} \\
			&  &&   FGM &  Carlini &  Anneal &  Decision \\
			Features & Classifier & Forgery Type &       &          &         &           \\
			\midrule
			CLBP & Linear & random &    - &        - &    0.36 &      0.60 \\
			&& skilled &    - &        - &    0.36 &      0.19 \\
			SigNet & Linear & random & 4.11 &     4.63 &       - &         - \\
			&& skilled & 4.26 &     2.01 &       - &      0.27 \\
			SigNet & RBF & random & 4.70 &     5.70 &       - &         - \\
			&& skilled & 4.19 &     2.45 &    3.92 &      1.30 \\
			
			\bottomrule
		\end{tabular}
	}
\end{table}

\begin{table}
	\centering
	\caption{Success rate of Type-I attacks on the Brazilian PUC-PR dataset (\% of attacks that transformed a genuine signature in a forgery)  (Limited Knowledge)}
	\label{tbl:success_evasion_lk1_brazilian}
	\begin{tabular}{ll|rrrr}
		\toprule
		& & \multicolumn{4}{c}{Attack Type} \\
		Feature &  Classifier &  FGM & Carlini & Anneal & Decision \\
		\midrule		
		CLBP & Linear   &      - &        - &   50.00 &     53.33 \\
		CLBP & RBF      &      - &        - &   90.00 &     90.00 \\
		SigNet & Linear & 100.00 &    80.00 &   50.00 &     43.33 \\
		SigNet & RBF    & 100.00 &    96.67 &   56.67 &     53.33 \\
		\bottomrule
	\end{tabular}
\end{table}

\begin{table}
	\centering
	\caption{Success rate of Type-II attacks on the Brazilian PUC-PR dataset (\% of attacks that transformed a forgery in a genuine signature)  (Limited Knowledge)}
	\label{tbl:success_intrusion_lk1_brazilian}
	\resizebox{\columnwidth}{!}{%
		\begin{tabular}{lll|rrrr}
			\toprule
			&&& \multicolumn{4}{c}{Attack Type} \\
			&  &&   FGM &  Carlini &  Anneal &  Decision \\
			Features & Classifier & Forgery Type &       &          &         &           \\
			\midrule
			CLBP & Linear & random &     - &        - &   26.67 &     33.33 \\
			&& skilled &     - &        - &   23.33 &     26.67 \\
			CLBP & RBF & random &     - &        - &    0.00 &      0.00 \\
			&& skilled &     - &        - &    0.00 &      0.00 \\
			SigNet & Linear & random &  0.00 &    50.00 &    0.00 &      0.00 \\
			&& skilled & 23.33 &    83.33 &    0.00 &      0.00 \\
			SigNet & RBF & random &  0.00 &    83.33 &    0.00 &      0.00 \\
			&& skilled & 10.00 &    93.33 &    0.00 &      0.00 \\			
			\bottomrule
		\end{tabular}
	}	
\end{table}

\begin{table}
	\centering
	\caption{Success rate of Type-I attacks on the Brazilian PUC-PR dataset (\% of attacks that transformed a genuine signature in a forgery)  (Limited Knowledge \#2)}
	\label{tbl:success_evasion_lk2_brazilian}
	\begin{tabular}{ll|rrrr}
		\toprule
		& & \multicolumn{4}{c}{Attack Type} \\
		Feature &  Classifier &  FGM & Carlini & Anneal & Decision \\
		\midrule
		SigNet & Linear & 80.00 &    10.00 &   43.33 &     13.33 \\
		SigNet & RBF    & 83.33 &    10.00 &   56.67 &     20.00 \\
		
		\bottomrule
	\end{tabular}
\end{table}

\begin{table}
	\centering
	\caption{Success rate of Type-II attacks on the Brazilian PUC-PR dataset (\% of attacks that transformed a forgery in a genuine signature)  (Limited Knowledge \#2)}
	\label{tbl:success_intrusion_lk2_brazilian}
	\resizebox{\columnwidth}{!}{%
		\begin{tabular}{lll|rrrr}
			\toprule
			&&& \multicolumn{4}{c}{Attack Type} \\
			&  &&   FGM &  Carlini &  Anneal &  Decision \\
			Features & Classifier & Forgery Type &       &          &         &           \\
			\midrule
			SigNet & Linear & random &  0.00 &     0.00 &    0.00 &      0.00 \\
			&& skilled & 10.00 &    10.00 &    0.00 &      0.00 \\
			SigNet & RBF & random &  0.00 &     0.00 &    0.00 &      0.00 \\
			&& skilled &  0.00 &     3.33 &    0.00 &      0.00 \\

			\bottomrule
		\end{tabular}
	}	
\end{table}

\begin{table}
	\caption{Success rate of Type-I attacks on the Brazilian PUC-PR dataset considering different defenses and attacker knowledge scenarios}
	\label{tbl:success_defenses_evasion_brazilian}
	\resizebox{\columnwidth}{!}{%
		\begin{tabular}{ll|rrr|rrr}
			\toprule
			&& \multicolumn{6}{c}{Attack Type and Knowledge scenario} \\
			&  &   \multicolumn{3}{c}{FGM} & \multicolumn{3}{c}{Carlini}\\
			Defense & Classifier & PK &   LK1 &   LK2 &     PK &   LK1 &  LK2 \\
			\midrule
			
			baseline & Linear & 100.00 & 100.00 & 80.00 &  100.00 &  86.67 & 10.00 \\
			& RBF & 100.00 & 100.00 & 83.33 &  100.00 &  96.67 & 10.00 \\
			ensadv & Linear & 100.00 &  96.67 & 76.67 &  100.00 &  80.00 &  3.33 \\
			& RBF & 100.00 & 100.00 & 83.33 &  100.00 &  96.67 &  3.33 \\
			madry & Linear & 100.00 &  93.33 & 33.33 &  100.00 &  83.33 &  3.33 \\
			& RBF & 100.00 & 100.00 & 36.67 &  100.00 & 100.00 &  0.00 \\
			
			\bottomrule
		\end{tabular}
	}
\end{table}

\begin{table}
	\caption{Distortion (RMSE of the adversarial noise) for Type-I attacks on the Brazilian PUC-PR dataset, considering different defenses and attacker knowledge scenarios}
	\label{tbl:mse_defenses_evasion_brazilian}
	\resizebox{\columnwidth}{!}{%
		\begin{tabular}{ll|rrr|rrr}
			\toprule
			&& \multicolumn{6}{c}{Attack Type and Knowledge scenario} \\
			&  &   \multicolumn{3}{c}{FGM} & \multicolumn{3}{c}{Carlini}\\
			Defense & Classifier & PK &   LK1 &   LK2 &     PK &   LK1 &  LK2 \\
			\midrule
			baseline & Linear & 4.09 & 4.06 & 4.24 &    0.96 & 0.97 & 1.18 \\
			& RBF & 4.09 & 4.00 & 4.11 &    1.03 & 1.04 & 1.27 \\
			ensadv & Linear & 4.28 & 4.26 & 4.15 &    0.94 & 0.92 & 1.51 \\
			& RBF & 4.31 & 4.35 & 4.12 &    0.98 & 0.97 & 1.59 \\
			madry & Linear & 4.57 & 4.54 & 4.39 &    2.16 & 2.35 & 0.83 \\
			& RBF & 4.57 & 4.64 & 4.18 &    2.41 & 2.55 &    - \\
			
			\bottomrule
		\end{tabular}
	}
\end{table}

\begin{table*}
	\centering
	\caption{Success rate of Type-II attacks on the Brazilian PUC-PR dataset considering different defenses and attacker knowledge scenarios}
	\label{tbl:success_defenses_intrusion_brazilian}
	
	\begin{tabular}{lll|rrr|rrr}
		\toprule
		&&& \multicolumn{6}{c}{Attack Type and Knowledge scenario} \\
		&&  &   \multicolumn{3}{c}{FGM} & \multicolumn{3}{c}{Carlini}\\
		Defense & Classifier & Forgery Type & PK &   LK1 &   LK2 &     PK &   LK1 &  LK2 \\
		\midrule
		baseline & Linear & random &  3.33 &  0.00 &  0.00 &  100.00 &  46.67 &  0.00 \\
		&              & skilled & 50.00 & 30.00 & 10.00 &  100.00 &  76.67 & 10.00 \\
		& RBF & random &  0.00 &  0.00 &  0.00 &  100.00 &  80.00 &  0.00 \\
		&              & skilled & 30.00 & 26.67 &  0.00 &  100.00 &  93.33 &  3.33 \\
		ensadv & Linear & random &  3.33 &  3.33 &  0.00 &  100.00 &  53.33 &  0.00 \\
		&              & skilled & 46.67 & 20.00 &  3.33 &  100.00 &  60.00 &  6.67 \\
		& RBF & random &  3.33 &  0.00 &  0.00 &  100.00 &  96.67 &  0.00 \\
		&              & skilled & 23.33 & 16.67 &  3.33 &  100.00 & 100.00 &  3.33 \\
		madry & Linear & random &  3.33 &  3.33 &  0.00 &   96.67 &  50.00 &  0.00 \\
		&              & skilled & 76.67 & 60.00 & 10.00 &  100.00 &  76.67 &  3.33 \\
		& RBF & random & 10.00 &  3.33 &  0.00 &   96.67 &  93.33 &  0.00 \\
		&              & skilled & 66.67 & 56.67 & 10.00 &  100.00 &  93.33 &  0.00 \\

		\bottomrule
	\end{tabular}
\end{table*}

\begin{table*}
	\centering
	\caption{Distortion (RMSE of the adversarial noise) for Type-II attacks on the Brazilian PUC-PR dataset, considering different defenses and attacker knowledge scenarios}
	\label{tbl:mse_defenses_intrusion_brazilian}
	\begin{tabular}{lll|rrr|rrr}
		\toprule
		&&& \multicolumn{6}{c}{Attack Type and Knowledge scenario} \\
		&&  &   \multicolumn{3}{c}{FGM} & \multicolumn{3}{c}{Carlini}\\
		Defense & Classifier & Forgery Type & PK &   LK1 &   LK2 &     PK &   LK1 &  LK2 \\
		\midrule
		baseline & Linear & random & 4.63 &    - &    - &    4.47 & 5.13 &    - \\
		&              & skilled & 4.25 & 4.06 & 4.21 &    1.86 & 1.79 & 2.05 \\
		& RBF & random &    - &    - &    - &    5.31 & 4.95 &    - \\
		&              & skilled & 4.01 & 3.85 &    - &    2.25 & 2.18 & 2.29 \\
		ensadv & Linear & random & 4.66 & 4.82 &    - &    6.83 & 7.46 &    - \\
		&              & skilled & 4.53 & 4.60 & 3.86 &    2.86 & 2.84 & 1.51 \\
		& RBF & random & 4.76 &    - &    - &    8.27 & 8.16 &    - \\
		&              & skilled & 4.62 & 4.60 & 3.84 &    3.89 & 3.69 & 2.45 \\
		madry & Linear & random & 5.12 & 5.38 &    - &    8.23 & 8.69 &    - \\
		&              & skilled & 4.89 & 4.86 & 3.95 &    3.88 & 3.69 & 1.30 \\
		& RBF & random & 4.79 & 5.28 &    - &    9.41 & 9.67 &    - \\
		&              & skilled & 4.86 & 4.74 & 4.27 &    4.96 & 4.73 &    - \\
		
		\bottomrule
	\end{tabular}
\end{table*}

\begin{table*}
	\centering
	\caption{Success of Type-I attacks on the Brazilian PUC-PR dataset in a PK scenario, with no pre-processing and with OTSU pre-processing}
	\label{tbl:otsu_evasion_brazilian}
	
	\begin{tabular}{ll|rrrrrrrr}
		\toprule
		&& \multicolumn{8}{c}{Attack Type and Preprocessing} \\
		& & \multicolumn{2}{c}{FGM} & \multicolumn{2}{c}{Carlini} & \multicolumn{2}{l}{Anneal} & \multicolumn{2}{c}{Decision} \\
		&  & None & OTSU & None & OTSU & None & OTSU & None & OTSU\\
		Feature & Classifier  &          &              &         &              &         &              &          &              \\
		\midrule
		CLBP & Linear &       - &            - &       - &            - &   66.67 &        13.33 &    76.67 &         3.33 \\
		& RBF &       - &            - &       - &            - &  100.00 &         0.00 &   100.00 &         0.00 \\
		SigNet Baseline & Linear &  100.00 &       100.00 &  100.00 &        10.00 &   90.00 &         0.00 &   100.00 &         6.67 \\
		& RBF &  100.00 &        96.67 &  100.00 &        23.33 &  100.00 &         0.00 &    96.67 &         0.00 \\
		SigNet Ens. Adv. & Linear &  100.00 &        93.33 &  100.00 &         6.67 &   96.67 &         0.00 &   100.00 &         0.00 \\
		& RBF &  100.00 &        93.33 &  100.00 &         3.33 &   96.67 &         0.00 &   100.00 &         0.00 \\
		SigNet Madry & Linear &  100.00 &       100.00 &  100.00 &        90.00 &   93.33 &         0.00 &   100.00 &         6.67 \\
		& RBF &  100.00 &       100.00 &  100.00 &        86.67 &   96.67 &         0.00 &   100.00 &         6.67 \\

		\bottomrule
	\end{tabular}
\end{table*}

\begin{table*}
	\centering
	\caption{Success of Type-II attacks on the Brazilian PUC-PR dataset in a PK scenario, with no pre-processing and with OTSU pre-processing}
	\label{tbl:otsu_intrusion_brazilian}
	\begin{tabular}{lll|rrrrrrrr}
		\toprule
		&&& \multicolumn{8}{c}{Attack Type and Preprocessing} \\
		&& & \multicolumn{2}{c}{FGM} & \multicolumn{2}{c}{Carlini} & \multicolumn{2}{l}{Anneal} & \multicolumn{2}{c}{Decision} \\
		&&  & None & OTSU & None & OTSU & None & OTSU & None & OTSU\\
		Feature & Classifier  & Forgery Type &         &              &         &              &         &              &          &              \\
		\midrule
		CLBP & Linear & random &       - &            - &       - &            - &   33.33 &         0.00 &    36.67 &         0.00 \\
		&              & skilled &       - &            - &       - &            - &   33.33 &         0.00 &    36.67 &         0.00 \\
		& RBF & random &       - &            - &       - &            - &    0.00 &         0.00 &     0.00 &         0.00 \\
		&              & skilled &       - &            - &       - &            - &    0.00 &         0.00 &     0.00 &         0.00 \\
\bottomrule
	\end{tabular}
\end{table*}

\clearpage

\begin{table}
	\centering
	\caption{Success rate of Type-I attacks on the GPDS-160 dataset(\% of attacks that transformed a genuine signature in a forgery)}
	\label{tbl:success_evasion_gpds}
	\begin{tabular}{ll|rrrr}
		\toprule
		& & \multicolumn{4}{c}{Attack Type} \\
		Feature &  Classifier &  FGM & Carlini & Anneal & Decision \\
		\midrule
		CLBP & Linear   &      - &        - &   62.82 &     83.33 \\
		CLBP & RBF      &      - &        - &  100.00 &    100.00 \\
		SigNet & Linear & 100.00 &   100.00 &   98.72 &    100.00 \\
		SigNet & RBF    & 100.00 &   100.00 &   98.72 &    100.00 \\

		\bottomrule
	\end{tabular}
\end{table}

\begin{table}
	\centering
	\caption{Distortion (RMSE of the adversarial noise) for successful Type-I attacks on the GPDS-160 dataset}
	\label{tbl:mse_evasion_gpds}
	\begin{tabular}{ll|rrrr}
		\toprule
		& & \multicolumn{4}{c}{Attack Type} \\
		Feature &  Classifier &  FGM & Carlini & Anneal & Decision \\
		\midrule
		
		CLBP & Linear   &    - &        - &    0.46 &      1.99 \\
		CLBP & RBF      &    - &        - &    0.36 &      $10^{-9}$ \\
		SigNet & Linear & 4.09 &     1.30 &    5.52 &      3.22 \\
		SigNet & RBF    & 4.13 &     1.36 &    5.12 &      3.00 \\

		\bottomrule
	\end{tabular}
\end{table}

\begin{table}
	\centering
	\caption{Success rate of Type-II attacks on the GPDS-160 dataset (\% of attacks that transformed a forgery in a genuine signature)}
	\label{tbl:success_intrusion_gpds}
	\resizebox{\columnwidth}{!}{%
		\begin{tabular}{lll|rrrr}
			\toprule
			&&& \multicolumn{4}{c}{Attack Type} \\
			&  &&   FGM &  Carlini &  Anneal &  Decision \\
			Features & Classifier & Forgery Type &       &          &         &           \\
			\midrule
			
			CLBP & Linear & random &     - &        - &   38.75 &     48.75 \\
			&& skilled &     - &        - &   38.75 &     48.75 \\
			CLBP & RBF & random &     - &        - &    0.00 &      0.00 \\
			&& skilled &     - &        - &    0.00 &      0.00 \\
			SigNet & Linear & random &  0.00 &    92.50 &    0.00 &      0.00 \\
			&& skilled & 21.25 &    98.75 &    2.50 &      2.50 \\
			SigNet & RBF & random &  0.00 &    88.75 &    0.00 &      0.00 \\
			&& skilled & 16.25 &   100.00 &    1.25 &      1.25 \\

			\bottomrule
		\end{tabular}
	}
	
\end{table}

\begin{table}
	\centering
	\caption{Distortion (RMSE of the adversarial noise) for successful Type-II attacks on the GPDS-160 dataset}
	\label{tbl:mse_intrusion_gpds}
	\resizebox{\columnwidth}{!}{%
		\begin{tabular}{lll|rrrr}
			\toprule
			&&& \multicolumn{4}{c}{Attack Type} \\
			&  &&   FGM &  Carlini &  Anneal &  Decision \\
			Features & Classifier & Forgery Type &       &          &         &           \\
			\midrule
			
			CLBP & Linear & random &    - &        - &    0.37 &      1.80 \\
			&& skilled &    - &        - &    0.39 &      1.51 \\
			SigNet & Linear & random &    - &     6.78 &       - &         - \\
			&& skilled & 4.18 &     3.54 &    5.71 &      2.74 \\
			SigNet & RBF & random &    - &     6.83 &       - &         - \\
			&& skilled & 4.11 &     3.98 &    4.66 &      1.96 \\
			
			\bottomrule
		\end{tabular}
	}
\end{table}

\begin{table}
	\centering
	\caption{Success rate of Type-I attacks on the GPDS-160 dataset (\% of attacks that transformed a genuine signature in a forgery)  (Limited Knowledge)}
	\label{tbl:success_evasion_lk1_gpds}
	\begin{tabular}{ll|rrrr}
		\toprule
		& & \multicolumn{4}{c}{Attack Type} \\
		Feature &  Classifier &  FGM & Carlini & Anneal & Decision \\
		\midrule		
		
		CLBP & Linear   &      - &        - &   44.87 &     44.87 \\
		CLBP & RBF      &      - &        - &   80.77 &     80.77 \\
		SigNet & Linear &  98.72 &    80.77 &   46.15 &     37.18 \\
		SigNet & RBF    & 100.00 &    91.03 &   50.00 &     44.87 \\

		\bottomrule
	\end{tabular}
\end{table}

\begin{table}
	\centering
	\caption{Success rate of Type-II attacks on the GPDS-160 dataset (\% of attacks that transformed a forgery in a genuine signature)  (Limited Knowledge)}
	\label{tbl:success_intrusion_lk1_gpds}
	\resizebox{\columnwidth}{!}{%
		\begin{tabular}{lll|rrrr}
			\toprule
			&&& \multicolumn{4}{c}{Attack Type} \\
			&  &&   FGM &  Carlini &  Anneal &  Decision \\
			Features & Classifier & Forgery Type &       &          &         &           \\
			\midrule
			
			CLBP & Linear & random &     - &        - &   23.75 &     28.75 \\
			&& skilled &     - &        - &   21.25 &     27.50 \\
			CLBP & RBF & random &     - &        - &    0.00 &      0.00 \\
			&& skilled &     - &        - &    0.00 &      0.00 \\
			SigNet & Linear & random &  0.00 &    45.00 &    0.00 &      0.00 \\
			&& skilled & 20.00 &    66.25 &    1.25 &      0.00 \\
			SigNet & RBF & random &  0.00 &    71.25 &    0.00 &      0.00 \\
			&& skilled & 18.75 &    80.00 &    0.00 &      0.00 \\

			\bottomrule
		\end{tabular}
	}	
\end{table}

\begin{table}
	\centering
	\caption{Success rate of Type-I attacks on the GPDS dataset (\% of attacks that transformed a genuine signature in a forgery)  (Limited Knowledge \#2)}
	\label{tbl:success_evasion_lk2_gpds}
	\begin{tabular}{ll|rrrr}
		\toprule
		& & \multicolumn{4}{c}{Attack Type} \\
		Feature &  Classifier &  FGM & Carlini & Anneal & Decision \\
		\midrule
		SigNet & Linear & 60.00 &     6.25 &   52.50 &     20.00 \\
		SigNet & RBF    & 63.75 &     7.50 &   51.25 &     17.50 \\
		
		\bottomrule
	\end{tabular}
\end{table}

\begin{table}
	\centering
	\caption{Success rate of Type-II attacks on the GPDS dataset (\% of attacks that transformed a forgery in a genuine signature)  (Limited Knowledge \#2)}
	\label{tbl:success_intrusion_lk2_gpds}
	\resizebox{\columnwidth}{!}{%
		\begin{tabular}{lll|rrrr}
			\toprule
			&&& \multicolumn{4}{c}{Attack Type} \\
			&  &&   FGM &  Carlini &  Anneal &  Decision \\
			Features & Classifier & Forgery Type &       &          &         &           \\
			\midrule
			SigNet & Linear & random & 0.00 &     0.00 &    0.00 &      0.00 \\
			&& skilled & 1.25 &     0.00 &    0.00 &      0.00 \\
			SigNet & RBF & random & 0.00 &     0.00 &    0.00 &      0.00 \\
			&& skilled & 1.25 &     0.00 &    0.00 &      0.00 \\

			\bottomrule
		\end{tabular}
	}	
\end{table}

\begin{table}
	\caption{Success rate of Type-I attacks on the GPDS dataset considering different defenses and attacker knowledge scenarios}
	\label{tbl:success_defenses_evasion_gpds}
	\resizebox{\columnwidth}{!}{%
		\begin{tabular}{ll|rrr|rrr}
			\toprule
			&& \multicolumn{6}{c}{Attack Type and Knowledge scenario} \\
			&  &   \multicolumn{3}{c}{FGM} & \multicolumn{3}{c}{Carlini}\\
			Defense & Classifier & PK &   LK1 &   LK2 &     PK &   LK1 &  LK2 \\
			\midrule
			baseline & Linear & 100.00 &  97.50 & 60.00 &  100.00 & 77.50 & 6.25 \\
			& RBF & 100.00 & 100.00 & 63.75 &  100.00 & 85.00 & 7.50 \\
			ensadv & Linear &  95.00 &  86.25 & 38.75 &  100.00 & 76.25 & 3.75 \\
			& RBF &  91.25 &  83.75 & 38.75 &  100.00 & 87.50 & 7.50 \\
			madry & Linear &  96.25 &  92.50 & 22.50 &  100.00 & 68.75 & 1.25 \\
			& RBF &  96.25 &  93.75 & 21.25 &  100.00 & 85.00 & 1.25 \\

			\bottomrule
		\end{tabular}
	}
\end{table}

\begin{table}
	\caption{Distortion (RMSE of the adversarial noise) for Type-I attacks on the GPDS dataset, considering different defenses and attacker knowledge scenarios}
	\label{tbl:mse_defenses_evasion_gpds}
	\resizebox{\columnwidth}{!}{%
		\begin{tabular}{ll|rrr|rrr}
			\toprule
			&& \multicolumn{6}{c}{Attack Type and Knowledge scenario} \\
			&  &   \multicolumn{3}{c}{FGM} & \multicolumn{3}{c}{Carlini}\\
			Defense & Classifier & PK &   LK1 &   LK2 &     PK &   LK1 &  LK2 \\
			\midrule
			baseline & Linear & 4.25 & 4.27 & 4.37 &    1.27 & 1.29 & 1.07 \\
			& RBF & 4.28 & 4.32 & 4.42 &    1.34 & 1.33 & 1.32 \\
			ensadv & Linear & 4.39 & 4.33 & 4.31 &    1.29 & 1.33 & 2.02 \\
			& RBF & 4.38 & 4.33 & 4.36 &    1.37 & 1.36 & 1.53 \\
			madry & Linear & 4.88 & 4.85 & 4.28 &    3.17 & 3.26 & 1.99 \\
			& RBF & 4.90 & 4.85 & 4.34 &    3.46 & 3.45 & 2.19 \\
			
			\bottomrule
		\end{tabular}
	}
\end{table}

\begin{table*}
	\centering
	\caption{Success rate of Type-II attacks on the GPDS dataset considering different defenses and attacker knowledge scenarios}
	\label{tbl:success_defenses_intrusion_gpds}
	
	\begin{tabular}{lll|rrr|rrr}
		\toprule
		&&& \multicolumn{6}{c}{Attack Type and Knowledge scenario} \\
		&&  &   \multicolumn{3}{c}{FGM} & \multicolumn{3}{c}{Carlini}\\
		Defense & Classifier & Forgery Type & PK &   LK1 &   LK2 &     PK &   LK1 &  LK2 \\
		\midrule
		baseline & Linear & random &  1.25 &  0.00 & 0.00 &   97.50 & 38.75 & 0.00 \\
		&              & skilled & 37.50 & 28.75 & 1.25 &  100.00 & 68.75 & 0.00 \\
		& RBF & random &  0.00 &  0.00 & 0.00 &   91.25 & 62.50 & 0.00 \\
		&              & skilled & 33.75 & 23.75 & 1.25 &  100.00 & 83.75 & 0.00 \\
		ensadv & Linear & random &  0.00 &  0.00 & 0.00 &   86.25 & 35.00 & 0.00 \\
		&              & skilled & 22.50 & 15.00 & 1.25 &  100.00 & 60.00 & 2.50 \\
		& RBF & random &  0.00 &  0.00 & 0.00 &   86.25 & 60.00 & 0.00 \\
		&              & skilled & 22.50 & 16.25 & 1.25 &  100.00 & 80.00 & 2.50 \\
		madry & Linear & random &  0.00 &  0.00 & 0.00 &   97.50 & 42.50 & 0.00 \\
		&              & skilled & 41.25 & 33.75 & 5.00 &  100.00 & 73.75 & 2.50 \\
		& RBF & random &  0.00 &  0.00 & 0.00 &   96.25 & 66.25 & 0.00 \\
		&              & skilled & 40.00 & 35.00 & 3.75 &  100.00 & 85.00 & 1.25 \\

		\bottomrule
	\end{tabular}
\end{table*}

\begin{table*}
	\centering
	\caption{Distortion (RMSE of the adversarial noise) for Type-II attacks on the GPDS dataset, considering different defenses and attacker knowledge scenarios}
	\label{tbl:mse_defenses_intrusion_gpds}
	\begin{tabular}{lll|rrr|rrr}
		\toprule
		&&& \multicolumn{6}{c}{Attack Type and Knowledge scenario} \\
		&&  &   \multicolumn{3}{c}{FGM} & \multicolumn{3}{c}{Carlini}\\
		Defense & Classifier & Forgery Type & PK &   LK1 &   LK2 &     PK &   LK1 &  LK2 \\
		\midrule
		baseline & Linear & random & 3.40 &    - &    - &    6.92 &  5.58 &    - \\
		&              & skilled & 4.17 & 4.12 & 4.32 &    3.40 &  2.78 &    - \\
		& RBF & random &    - &    - &    - &    6.64 &  6.06 &    - \\
		&              & skilled & 4.07 & 3.99 & 4.41 &    3.60 &  3.32 &    - \\
		ensadv & Linear & random &    - &    - &    - &    8.88 &  9.24 &    - \\
		&              & skilled & 4.59 & 4.61 & 4.32 &    4.98 &  3.89 & 1.67 \\
		& RBF & random &    - &    - &    - &    9.48 &  9.33 &    - \\
		&              & skilled & 4.66 & 4.66 & 4.31 &    5.54 &  4.74 & 2.14 \\
		madry & Linear & random &    - &    - &    - &   11.55 & 10.81 &    - \\
		&              & skilled & 4.90 & 4.92 & 4.10 &    6.62 &  5.87 & 1.99 \\
		& RBF & random &    - &    - &    - &   11.98 & 11.97 &    - \\
		&              & skilled & 4.95 & 4.91 & 3.98 &    7.22 &  6.63 & 1.67 \\
		
		\bottomrule
	\end{tabular}
\end{table*}

\begin{table*}
	\centering
	\caption{Success of Type-I attacks on the GPDS dataset in a PK scenario, with no pre-processing and with OTSU pre-processing}
	\label{tbl:otsu_evasion_gpds}
	
	\begin{tabular}{ll|rrrrrrrr}
		\toprule
		&& \multicolumn{8}{c}{Attack Type and Preprocessing} \\
		& & \multicolumn{2}{c}{FGM} & \multicolumn{2}{c}{Carlini} & \multicolumn{2}{l}{Anneal} & \multicolumn{2}{c}{Decision} \\
		&  & None & OTSU & None & OTSU & None & OTSU & None & OTSU\\
		Feature & Classifier  &          &              &         &              &         &              &          &              \\
		\midrule
		CLBP & Linear &       - &            - &       - &            - &   62.82 &         8.97 &    83.33 &         5.13 \\
		& RBF &       - &            - &       - &            - &  100.00 &         1.28 &   100.00 &         0.00 \\
		SigNet Baseline & Linear &  100.00 &        90.00 &  100.00 &        22.50 &   98.75 &         1.25 &   100.00 &         1.25 \\
		& RBF &  100.00 &        88.75 &  100.00 &        23.75 &   98.75 &         0.00 &    98.75 &         1.25 \\
		SigNet Ens. Adv. & Linear &   95.00 &        63.75 &  100.00 &         3.75 &  100.00 &         0.00 &   100.00 &         0.00 \\
		& RBF &   91.25 &        63.75 &  100.00 &         2.50 &  100.00 &         0.00 &   100.00 &         0.00 \\
		SigNet Madry & Linear &   96.25 &        92.50 &  100.00 &        78.75 &   92.50 &         0.00 &   100.00 &        10.00 \\
		& RBF &   96.25 &        95.00 &  100.00 &        77.50 &   91.25 &         0.00 &   100.00 &         8.75 \\

		\bottomrule
	\end{tabular}
\end{table*}

\begin{table*}
	\centering
	\caption{Success of Type-II attacks on the GPDS dataset in a PK scenario, with no pre-processing and with OTSU pre-processing}
	\label{tbl:otsu_intrusion_gpds}
	\begin{tabular}{lll|rrrrrrrr}
		\toprule
		&&& \multicolumn{8}{c}{Attack Type and Preprocessing} \\
		&& & \multicolumn{2}{c}{FGM} & \multicolumn{2}{c}{Carlini} & \multicolumn{2}{l}{Anneal} & \multicolumn{2}{c}{Decision} \\
		&&  & None & OTSU & None & OTSU & None & OTSU & None & OTSU\\
		Feature & Classifier  & Forgery Type &         &              &         &              &         &              &          &              \\
		\midrule
		CLBP & Linear & random &       - &            - &       - &            - &   38.75 &         1.25 &    48.75 &         0.00 \\
		&              & skilled &       - &            - &       - &            - &   38.75 &         3.75 &    48.75 &         0.00 \\
		& RBF & random &       - &            - &       - &            - &    0.00 &         0.00 &     0.00 &         0.00 \\
		&              & skilled &       - &            - &       - &            - &    0.00 &         0.00 &     0.00 &         0.00 \\
\bottomrule
	\end{tabular}
\end{table*}
\clearpage


\end{document}